%% file: main.tex
\documentclass{article}
\usepackage[preprint]{neurips_2026}

\usepackage{todonotes}

\usepackage{array}
\usepackage[T1]{fontenc}
\usepackage[utf8]{inputenc}
\usepackage{graphicx}
\usepackage{booktabs}
\usepackage{amsmath}
\usepackage{amssymb}
\usepackage{amsfonts}
\usepackage{multirow}
\usepackage{verbatim}
\usepackage{caption}
\usepackage{longtable}
\usepackage{supertabular}
\usepackage{float}
\usepackage{mathtools}
\usepackage{latexsym}

\usepackage{enumitem}
\usepackage{tablefootnote}
\usepackage{xcolor}
\usepackage{xspace}
\usepackage{textcomp}
\usepackage{makecell}
\usepackage{lscape} 
\usepackage{siunitx}
\usepackage{amssymb}
\usepackage{bm}
\usepackage{algorithm}
\usepackage{algpseudocode}
\usepackage{graphicx}
\usepackage{wrapfig}
\usepackage{caption}

\newcommand{\cmark}{\ding{51}}%
\newcommand{\xmark}{\ding{55}}%
\usepackage{scrextend}

\usepackage{microtype}
\definecolor{mydarkblue}{rgb}{0,0.08,0.45}
\usepackage[colorlinks,citecolor=mydarkblue,urlcolor=mydarkblue,linkcolor=mydarkblue]{hyperref}
\usepackage{url}            %
\usepackage{nicefrac}       %
\usepackage{changepage}
\usepackage{xargs}          %
\usepackage{wrapfig,lipsum,booktabs}

\usepackage{subcaption}
\usepackage{endnotes}

\usepackage{pgfplots}
\usetikzlibrary{pgfplots.groupplots}
\pgfplotsset{compat=1.3}
\usepackage{tikz}
\usetikzlibrary{patterns}
\usepackage{sidecap}
\usepackage{pifont}

\usepackage[capitalize,noabbrev]{cleveref}
\crefname{section}{Section}{Sections}
\Crefname{section}{Section}{Sections}
\crefname{table}{Table}{Tables}
\crefname{figure}{Figure}{Figures}
\crefname{algorithm}{Algorithm}{Algorithms}
\crefname{equation}{eq.}{eqs.}
\crefname{appendix}{Appendix}{Appendices}
\crefformat{section}{Section #2#1#3}
\usepackage{multicol}

\usepackage{minitoc}
\usepackage{listings}
\usepackage{colortbl}

\title{SpatialOPSD: Self-Distilling Spatial Intelligence from Verified Coding Agent Traces}

\author{%
Rongxue Li\thanks{This work was done during the author’s internship.}\\
Alibaba Group\\
\And
Meng Yang\footnotemark[1] \\
Alibaba Group\\
\And
Yiru Mao \\
Alibaba Group\\
\And
Yongliang Tao \\
Alibaba Group\\
\And
Lulu Hu\\
Alibaba Group\\
\And
Bin Yang\\
Alibaba Group\\
\And
Zhao Xu\\
Alibaba Group\\
\And
Weihua Luo\\
Alibaba Group\\
\And
Bowen Xu\thanks{Corresponding author. Email: bowen.xbw@alibaba-inc.com}\\
Alibaba Group\\
}

\usepackage{dsfont}
\input{math}
\input{listings_style}

\crefname{listing}{algorithm}{algorithms}
\Crefname{listing}{Algorithm}{Algorithms}

\definecolor{bestcolor}{RGB}{219, 208, 237}
\definecolor{secondcolor}{RGB}{241, 237, 248}
\definecolor{thirdcolor}{RGB}{211, 222, 190}
\definecolor{line-blue}{RGB}{243, 248, 252}
\definecolor{line-green}{RGB}{200,242,200}
\definecolor{line-red}{RGB}{255,215,215}

\begin{document}

\doparttoc %
\faketableofcontents %

\maketitle

\input{sections/0.Abstract}

\input{sections/1.Introduction}

\input{sections/3.Preliminary}

\input{sections/4.Method}

\input{sections/5.Experiments}

\input{sections/6.Analysis}

\input{sections/2.Related}
\input{sections/7.Conclusion}

% \clearpage

\bibliography{main}
\bibliographystyle{plainnat}

\clearpage
\appendix
\input{appendix/appendix}

% \clearpage
% \input{checklist}

\end{document}

%% file: listings_style.tex
\definecolor{codegreen}{rgb}{0,0.6,0}
\definecolor{codegray}{rgb}{0.5,0.5,0.5}
\definecolor{codepurple}{rgb}{0.58,0,0.82}
\definecolor{backcolour}{rgb}{0.95,0.95,0.92}

\lstdefinestyle{mystyle}{
  backgroundcolor=\color{backcolour}, commentstyle=\color{codegreen},
  keywordstyle=\color{magenta},
  numberstyle=\tiny\color{codegray},
  stringstyle=\color{codepurple},
  basicstyle=\ttfamily\footnotesize,
  breakatwhitespace=false,         
  breaklines=true,                 
  captionpos=b,                    
  keepspaces=true,                 
  numbers=left,                    
  numbersep=5pt,                  
  showspaces=false,                
  showstringspaces=false,
  showtabs=false,                  
  tabsize=2
}

%% file: sections/0.Abstract.tex
\begin{abstract}
Spatial coding agents significantly improve spatial reasoning in  Multimodal Large Language Models (MLLMs) by using external tools to generate verified execution traces. However, this paradigm inherently suffers from prohibitive inference-time overhead and external dependencies. In this paper, we explore whether an MLLM can internalize this agentic capability to operate entirely tool-free. We begin with a simple observation: prompting an MLLM with summarized execution traces of a spatial coding agent naturally unlocks the model's internal spatial Chain-of-Thought (CoT). Motivated by this, we introduce SpatialOPSD, an on-policy self-distillation framework that internalizes spatial reasoning into a standalone MLLM by formulating verified agent traces as privileged information. 
To mitigate privileged-information leakage during distillation, we introduce Repetition-Aware Distillation, which combines repetition masking with unlikelihood regularization.
Experiments across multiple benchmarks demonstrate that self-distilling SpatialOPSD achieves higher average accuracy than SFT and GRPO on both spatial and OOD datasets, exhibiting superior performance and 
generalization.

\end{abstract}

%% file: sections/1.Introduction.tex
% \vspace{-6pt}
\section{Introduction}
% \vspace{-6pt}

Spatial reasoning requires inferring latent 3D properties from 2D observations~\citep{yang2025thinking}, posing a fundamental challenge for multimodal large language models (MLLMs)~\citep{singh2025openai, qwen3.5}.
While final answers may be verifiable, the intermediate geometric inferences are difficult to annotate or validate automatically. 
This unverifiability bottlenecks standard training paradigms. Answer-level supervised fine-tuning (SFT) forces spatial understanding to remain implicit and brittle~\citep{daxberger2025mm, cai2026scaling, gao2026holi, liu2026openspatial, wu2026spatialscore}, while explicit Chain-of-Thought (CoT) supervision is prohibitively expensive to scale~\citep{wei2022chain, cai2026scaling}. Although Reinforcement Learning (e.g., GRPO)~\citep{shao2024deepseekmath} could bypass this annotation bottleneck via exploration, its application to spatial reasoning remains inefficient. Lacking verifiable intermediate steps, current RL pipelines are forced to rely on costly cold-start data and sparse outcome rewards~\citep{ sarch2026grounded, sun2025spacevista, li2026prosr}.

Motivated by the widespread success of coding agents, recent methods cast spatial reasoning as an agentic task~\citep{luo2026pyspatial, cho2026spatialclaw}. By equipping MLLMs with a code-execution interface and external geometry tools~\citep{lin2025depth, carion2026sam}, this paradigm elegantly resolves the aforementioned verifiability bottleneck. Yet, this solution merely shifts the burden to inference time. The iterative loop of decoding and tool execution introduces high latency and infrastructure overhead. This motivates a simple question: \textit{Can an MLLM internalize spatial reasoning from these verified agent traces, operating entirely tool-free at inference time?}

% \begingroup
\setlength{\columnsep}{8pt}
\setlength{\intextsep}{4pt}

\begin{wrapfigure}{r}{0.4\textwidth}
    \vspace{-\intextsep}
    \centering

    \captionsetup{
        font=small,
        skip=2pt
    }
    \captionsetup[subfigure]{
        font=footnotesize,
        skip=1pt
    }

    \begin{subfigure}[t]{\linewidth}
        \centering
        \includegraphics[
            width=\linewidth,
            height=3.2cm,
            keepaspectratio,
            trim=2pt 2pt 2pt 2pt,
            clip
        ]{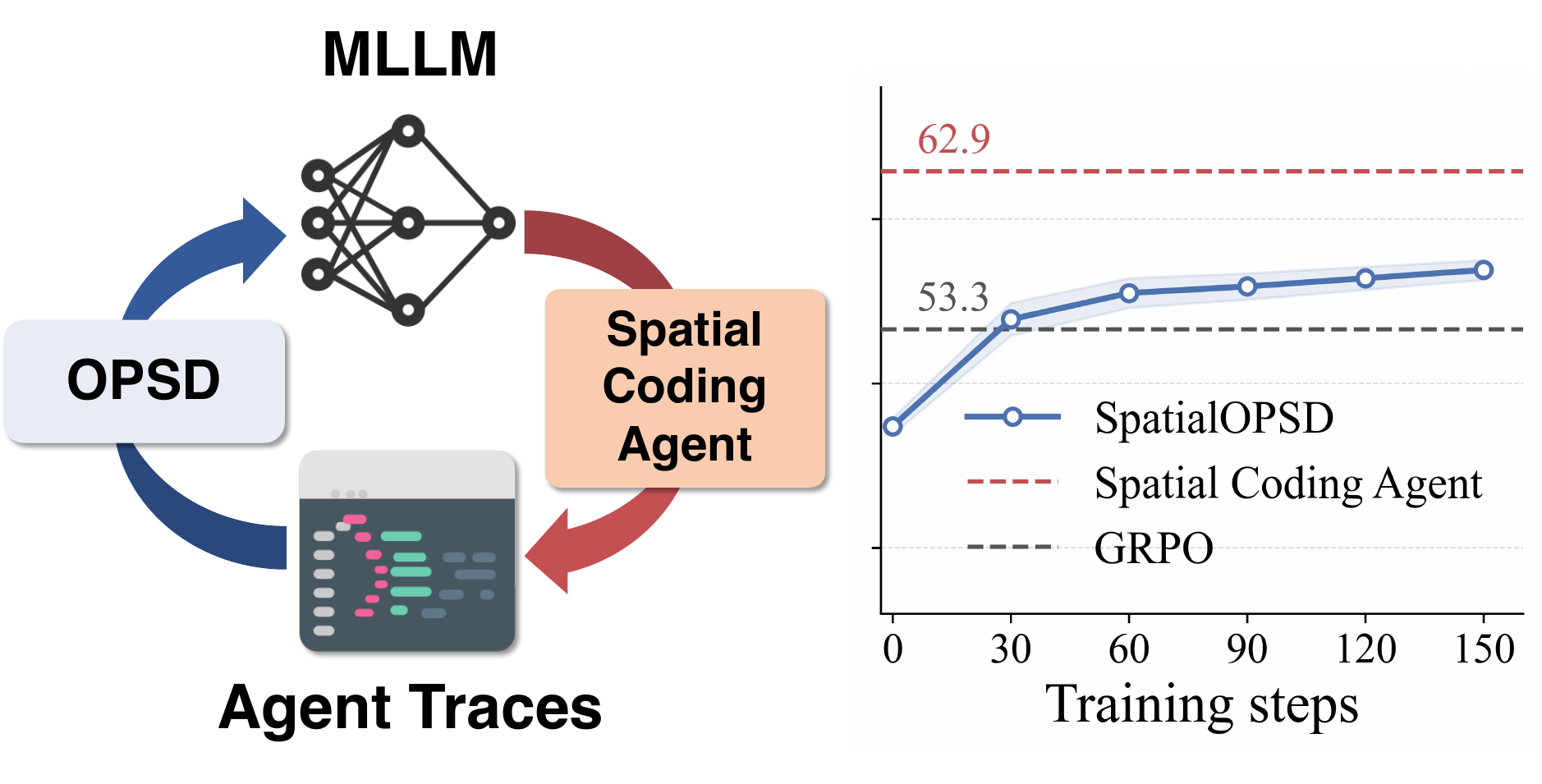}
        % \caption{Prompting strategies}
        \label{fig:intro}
    \end{subfigure}

    \caption{\textbf{MindCube-Tiny test accuracy}. OPSD from spatial agents effectively boosts spatial reasoning.}
    \label{fig:introduction}
    % \vspace{-4pt}
\end{wrapfigure}
To this end, we introduce SpatialOPSD, a framework that transfers tool-based spatial reasoning to MLLMs through on-policy self-distillation (OPSD)~\citep{zhao2026self}, without requiring tools at
inference time~(Figure~\ref{fig:introduction}). Our starting observation is that conditioning an MLLM on spatial coding agent traces elicits higher-quality CoT reasoning on spatial tasks. SpatialOPSD converts this inference-time benefit into training supervision by summarizing verified agent traces into privileged information for OPSD. 
To isolate spatial reasoning from tool-specific artifacts inherently coupled in execution traces, we pair this trace-privileged setup with \emph{Repetition-Aware Distillation}, which penalizes verbatim copying of privileged content and self-repetition.
Empirical results on MindCube~\citep{yin2025spatial} and ViewSpatial~\citep{li2026viewspatial} demonstrate that SpatialOPSD substantially surpasses standard SFT and RL baselines. It approaches the performance of tool-using spatial agents while operating entirely tool-free. It largely preserves performance on out-of-distribution (OOD)
benchmarks and yields gains across the evaluated model sizes, demonstrating its broad applicability. Finally, relying on self-generated traces, this paradigm establishes a self-distillation loop for geometric reasoning. 
% \endgroup

Our contributions are threefold:
\begin{itemize}[leftmargin=*, topsep=2pt, itemsep=1pt, parsep=0pt, partopsep=0pt]
\item We show that spatial coding-agent traces effectively elicit multimodal spatial reasoning.

\item We propose SpatialOPSD, which distills tool-augmented spatial reasoning into multimodal models.

\item We introduce Repetition-Aware Distillation to reduce privileged-information leakage during OPSD.
\end{itemize}

%% file: sections/3.Preliminary.tex
% \vspace{-10pt}
\section{Preliminary}
% \vspace{-10pt}
% \subsection{Coding Agent} 
\textbf{Spatial Coding Agent.}
A spatial coding agent~\citep{cho2026spatialclaw} operates over a persistent Python kernel equipped with perception modules and scientific libraries~\citep{lin2025depth, carion2026sam}. Given an input $x$, the agent iteratively writes and executes code to produce a $T$-step trace:
\begin{equation}
    \label{eq:agent_trace}
    \begin{aligned}
    \tau = (r_0, a_0, o_0 \dots r_t, a_t, o_t\dots, y_\mathcal{A}),
    \end{aligned}
\end{equation}
where $r_t$ is a natural-language reasoning step, $a_t$ is an executable code snippet, $o_t$ is the execution output, and \(y_\mathcal{A}\)
is the agent's final answer. At each step $t$, an MLLM policy $\pi_\theta$ generates the reasoning and code conditioned on the context history: $(r_t, a_t) \sim \pi_{\theta}(\cdot|x, \tau_{<t})$. The kernel then executes $a_t$ to yield $o_t$, while maintaining the state variables in Python kernel across turns for subsequent steps.

\textbf{On-Policy Self-Distillation.} On-policy self-distillation (OPSD) \citep{zhao2026self} uses a copy of the student's initial model, conditioned on privileged information $\mathbf{I}$, as its teacher.
Let $p_\theta$ denote the trainable student and $p_\phi$
the frozen teacher, both initialized from the same checkpoint.
For a student-generated trajectory $y$, the iteration-specific
training objective is
\begin{equation}
    \label{eq:opsd-trajectory}
    \ell_{\mathrm{OPSD}}(y)
    \triangleq
    \frac{1}{|y|}
    \sum_{t=1}^{|y|}
    D_{\mathrm{KL}}\!\left(
        \underbrace{
            p_\theta(\cdot \mid x, y_{<t})
        }_{\text{student}}
        \;\big\|\;
        \underbrace{
                p_\phi(\cdot \mid x, \mathbf{I}, y_{<t})
        }_{\text{teacher}}
    \right).
\end{equation}
Averaging over on-policy student rollouts yields the training loss:
\begin{equation}
    \label{eq:opsd}
    \mathcal{L}_{\mathrm{OPSD}}
    =
    \mathbb{E}_{y \sim p_{\bar{\theta}}(\cdot \mid x)}
    \left[\ell_{\mathrm{OPSD}}(y)\right],
    \qquad
    \bar{\theta} = \operatorname{sg}(\theta),
\end{equation}
where $\operatorname{sg}$ denotes stop-gradient.
Rollouts are sampled with detached parameters $\bar{\theta}$ and held
fixed during optimization; gradients flow only through the student branch. OPSD provides dense supervision on the student's own trajectories,
but introduces two risks:
\emph{self-reinforcing repetition}, where the teacher endorses
repetitive continuations \citep{holtzman2019curious, fang2026copy}; and
\emph{privileged-information leakage}, 
% where the teacher's
% context-copying bias encourages reproducing privileged content
% rather than learning task-relevant reasoning \citep{yang2026self}.
where the teacher's
context-copying bias may encourage the student to reproduce
details from privileged information that will be unavailable
at inference time, rather than learn task-relevant reasoning \citep{yang2026self}.

\textbf{Chain-of-Thought.} CoT introduces an intermediate rationale $z$ to expand test-time computation~\citep{wei2022chain}. It factorizes the prediction for input $x$ and answer $y$ as
\begin{equation}
p(y,z\mid x) = p(y\mid x, z) \cdot p(z \mid x).
\label{eq:cot}
\end{equation}
% Despite its profound success in general domains, text-based CoT is largely ineffective for spatial reasoning in MLLMs, and can even degrade accuracy
Despite its success in general reasoning tasks, text-based CoT
has shown mixed effects on spatial reasoning in MLLMs, improving
accuracy for some models and tasks while degrading it for
others
~\citep{yang2026mmsi,yin2025spatial,yang2025thinking}. Generating an intermediate rationale alone is therefore not
sufficient to ensure effective spatial reasoning.

%% file: sections/4.Method.tex
\section{Spatial Reasoning via OPSD}
\subsection{Motivation}

% \noindent
We ask whether execution-grounded guidance can elicit more effective spatial reasoning than a generic chain-of-thought prompt. To investigate this question, we compare four inference settings on shared spatial reasoning benchmarks~\citep{yin2025spatial, zhang2026flatland} using Qwen3.5-9B~\citep{qwen3.5}. We compare four settings on the same benchmarks: \textbf{Direct QA} answers without a rationale; \textbf{Prompted CoT} elicits step-by-step reasoning; \textbf{Spatial Coding Agent} iteratively generates and executes code; and \textbf{Agent-Trace-Conditioned CoT} uses the agent’s execution trace to produce a rationale and answer without further tool calls. Figure~\ref{fig:motivation_a} reveals an intriguing phenomenon: while deploying a coding agent significantly outperforms generic CoT prompting, merely conditioning CoT on observed agent traces matches the spatial reasoning capabilities of the full agent. As shown in Figure~\ref{fig:motivation_b}, these traces ground the generated rationales in concrete spatial analysis, significantly increasing references to 3D perception ($1.8\times$) and counterfactual motion ($1.6\times$) concepts. 

\begingroup
% \setlength{\columnsep}{10pt} % 图片与左侧正文之间的水平距离
% \setlength{\intextsep}{2pt} % 图片与上下正文之间的距离

% \begin{wrapfigure}[11]{r}{0.6\textwidth}
\begin{wrapfigure}{r}{0.6\textwidth}
    % \vspace{-6pt}
    \centering

    \captionsetup{
        font=small,
        skip=2pt
    }
    \captionsetup[subfigure]{
        font=footnotesize,
        skip=1pt
    }

    \begin{subfigure}[t]{0.48\linewidth}
        \centering
        \includegraphics[
            height=3.2cm,
            width=\linewidth,
            keepaspectratio,
            trim=2pt 2pt 2pt 2pt,
            clip
        ]{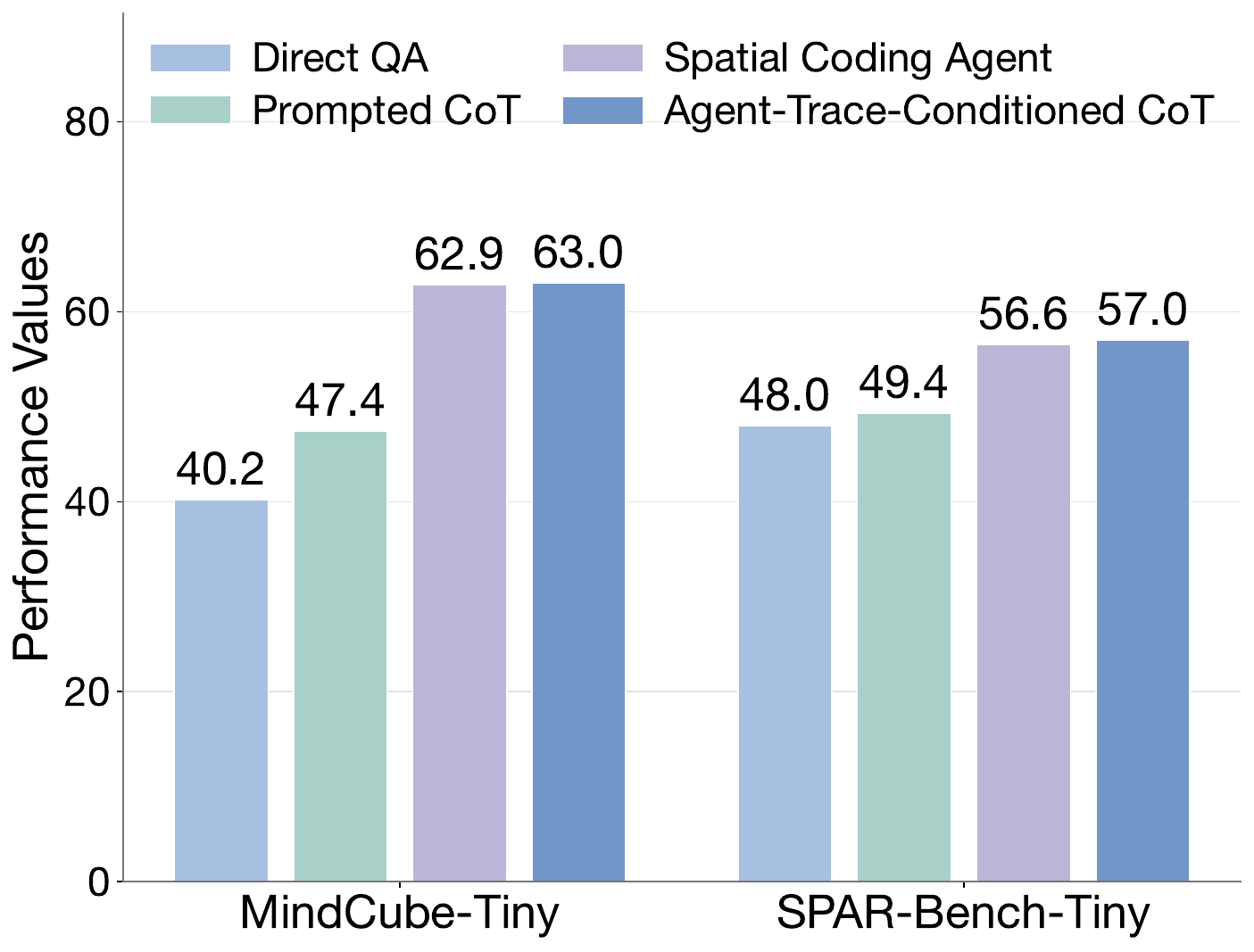}
        \caption{Prompting strategies}
        \label{fig:motivation_a}
    \end{subfigure}
    \hfill
    \begin{subfigure}[t]{0.48\linewidth}
        \centering
        \includegraphics[
            height=3.2cm,
            width=\linewidth,
            keepaspectratio,
            trim=2pt 2pt 2pt 2pt,
            clip
        ]{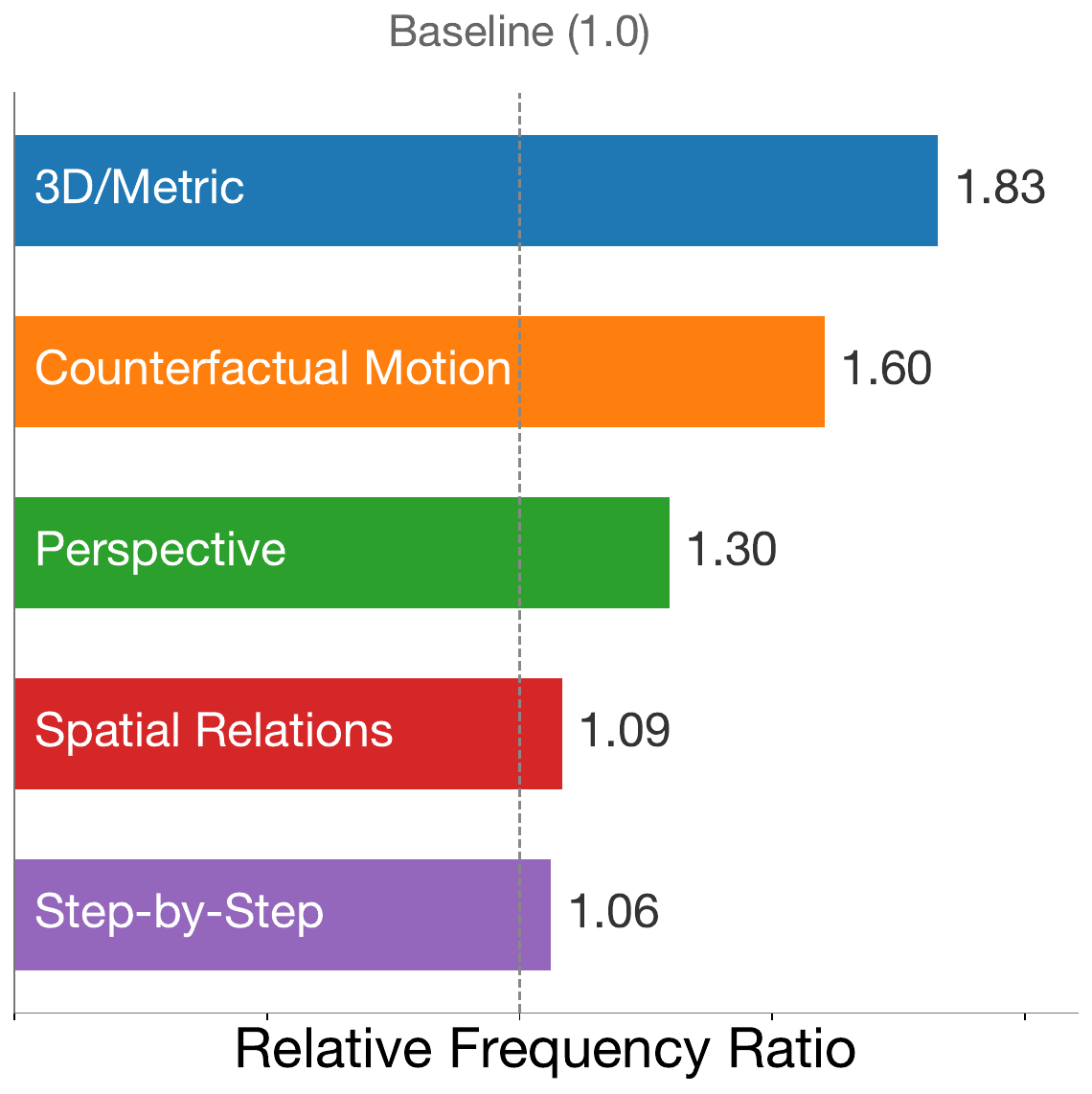}
        \caption{Frequency changes}
        \label{fig:motivation_b}
    \end{subfigure}
    \caption{\textbf{(a)} Effects of prompting strategies on spatial reasoning.  \textbf{(b)} Agent-trace-conditioned CoT increases spatial concept frequency.}
    \label{fig:motivation}
    % \vspace{-0pt}
\end{wrapfigure}
However, this paradigm still fundamentally relies on externally generated traces during inference. This motivates our core objective: \emph{can we internalize these trace-conditioned benefits to enable fully tool-free spatial reasoning?} We address this by formulating agent traces as privileged information during training. By distilling a teacher model conditioned on these traces into a student conditioned solely on the raw input, we compel the model to internalize the underlying geometric logic, rather than merely memorizing trace-specific artifacts.
\par
\endgroup

\subsection{SpatialOPSD}

\begin{figure}[!h]
    \centering
    \includegraphics[width=1.0\linewidth]{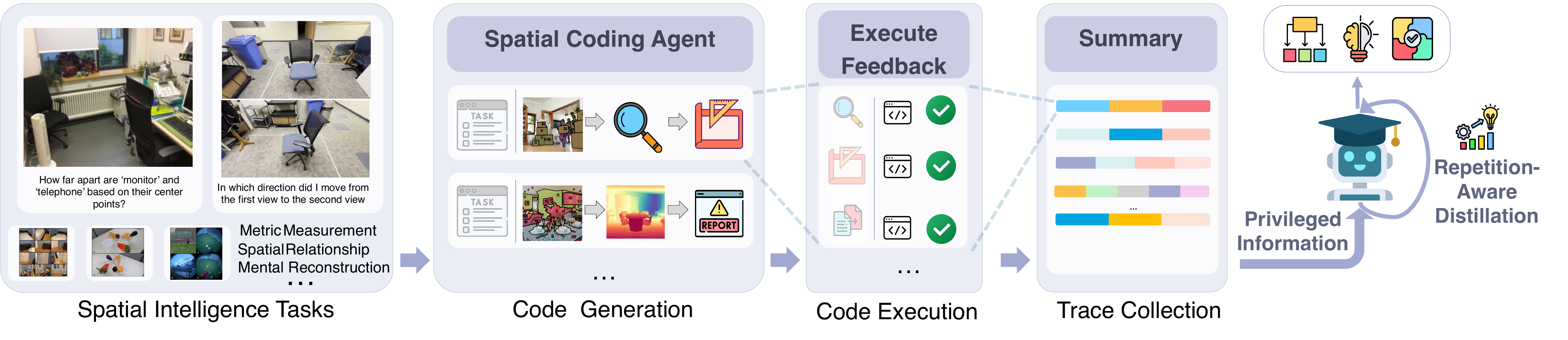}
    % \vspace{-5mm}
    \caption{Overview of SpatialOPSD. Verified spatial-agent traces are summarized into privileged information, which guides Repetition-Aware Distillation for spatial reasoning.
}
    \label{fig:d_core}
    % \vspace{-10pt}
\end{figure}

% \vspace{-6pt}
% \vspace{-6pt}
% Central to this formulation are two key designs. The first is defining \emph{what} to transfer. Because raw traces entangle rich geometric reasoning with tool-specific artifacts, we construct privileged representations that cleanly decouple essential spatial knowledge from the execution process. The second addresses a fundamental optimization challenge in OPSD: teachers tend to reinforce, rather than correct, a student's repetitive predictions. We introduce \emph{Repetition-Aware Distillation} to explicitly mask and penalize these degenerate loops. Together, this decoupled representation and robust objective enable the student to master tool-backed reasoning while remaining strictly tool-free at inference.
We present SpatialOPSD, an OPSD framework that learns from privileged execution traces. Our setup is simple: a single MLLM serves as both teacher and student, with asymmetric access to information. The teacher uses verified traces to provide token-level supervision; the student observes only the raw input $x$. To instantiate this framework, our first step is to acquire the privileged training context.

% \vspace{-2pt}
\textbf{Trace Collection.} We deploy a spatial coding agent $\mathcal{A}$ (e.g., SpatialClaw) as an oracle that solves each training query $x$ through a multi-turn code-execution loop. At turn $t$, $\mathcal{A}$ emits a reasoning step $r_t$, generates a code cell $a_t$, and receives a verified observation $o_t = \mathrm{exec}(a_t, s_t)$ from a persistent Python kernel equipped with geometric primitives (depth, segmentation, camera pose). A trace is the full interaction record $\tau = (r_0, a_0, o_0, \ldots, r_{T-1}, a_{T-1}, o_{T-1}, y_\mathcal{A})$, retained only when the final prediction $y_\mathcal{A}$ passes deterministic verification against the ground truth. This yields a corpus $\mathcal{D}_\tau = \{(x_i, \tau_i, y_i)\}$ of verified, load-bearing reasoning traces, without any human annotation.

% \vspace{-2pt}
\textbf{Privileged Information Construction.} Given a verified trace $\tau$ and answer $y$, we construct
privileged information $\mathbf{I}=\Phi(\tau,y)$ to condition
the teacher. We consider four variants:
\begin{equation}
\label{eq:priv_ops}
\Phi(\tau,y)=
\begin{cases}
\tau,
& \text{\textsc{full-trace}}, \\[4pt]
\mathrm{scrub}_y\!\left(g(\tau;\rho_{\mathrm{sum}})\right),
& \text{\textsc{summary}}, \\[4pt]
g(\tau;\rho_{\mathrm{fact}})\,\Vert\,\alpha(y),
& \text{\textsc{fact}}, \\[4pt]
\psi\!\left(g(\tau;\rho_{\mathrm{int}})\right),
& \text{\textsc{intent}}.
\end{cases}
\end{equation}
Here, $g(\tau;\rho_k)$ rewrites $\tau$ under instruction $\rho_k$.
\textsc{Full-trace} retains $\tau$ verbatim.
\textsc{Summary} condenses $(r_t,a_t,o_t)$, with
$\mathrm{scrub}_y$ removing explicit answer references.
\textsc{Fact} extracts factual findings without inferential
steps and appends the formatted answer $\alpha(y)$ via
concatenation $\Vert$.
\textsc{Intent} extracts the strategy underlying $(r_t,a_t)$,
with $\psi$ removing tool-specific details. Figure~\ref{fig:privileged information type} shows that \textsc{summary} performs best with a moderate teacher–student entropy gap, while other strategies exhibit high uncertainty or excessive entropy gaps; prompts and implementation details are provided in Appendix~\ref{app:priv-construction}.

\setlength{\columnsep}{8pt}
\setlength{\intextsep}{4pt}
\begin{wrapfigure}{r}{0.45\textwidth}
    \vspace{-\intextsep}
    \centering
    \captionsetup{
        font=small,
        skip=2pt
    }

    \includegraphics[
        width=\linewidth,
        trim={4pt 4pt 4pt 4pt},
        clip
    ]{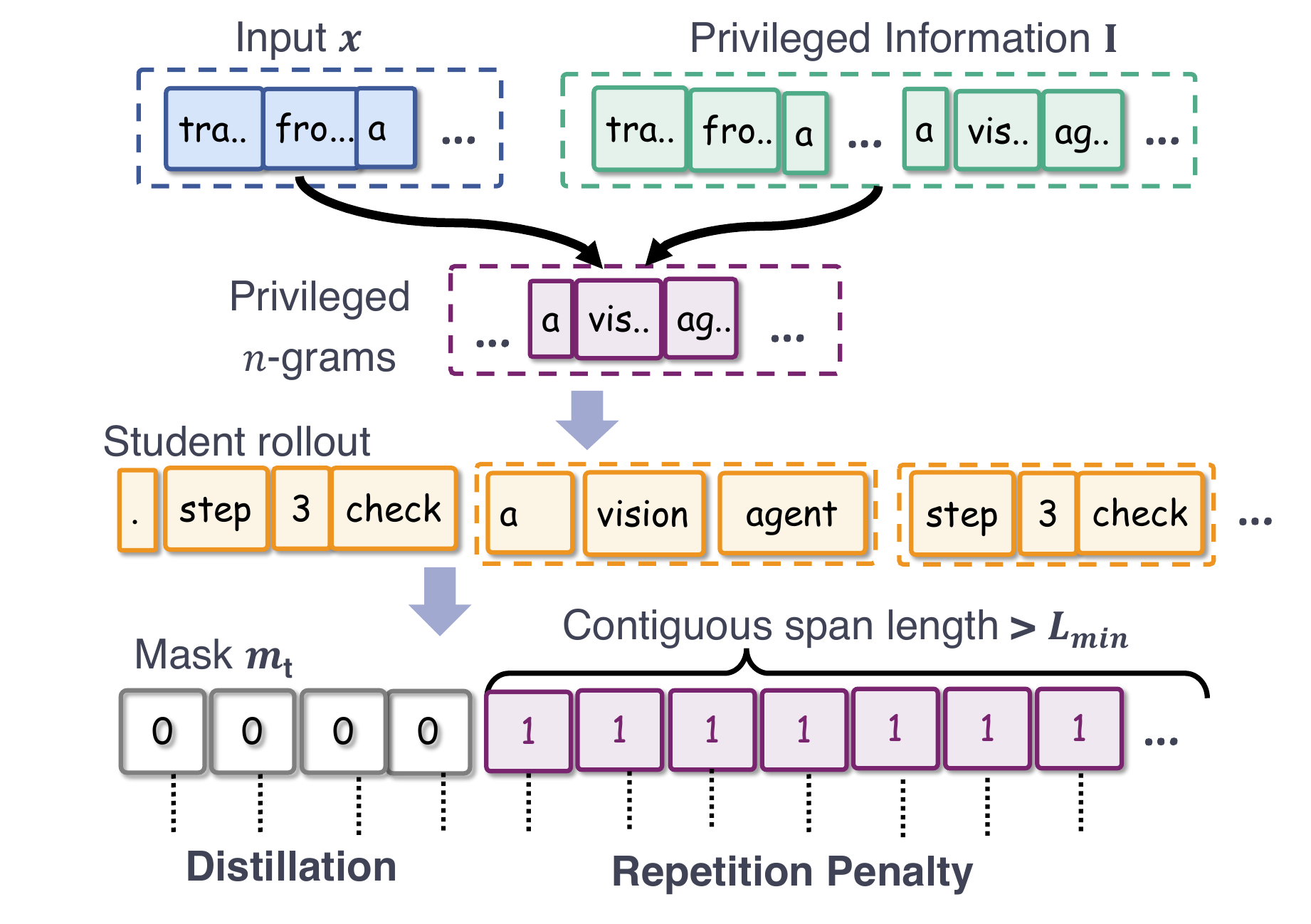}

    \caption{Illustration of Repetition-Aware Distillation.}
    \label{fig:repetition_distillation}

    % \vspace{-4pt}
\end{wrapfigure}
% 这里直接接需要环绕的正文，不要在环境和正文之间插入空行。

\textbf{Repetition-Aware Distillation.} 
Given the privileged information $\mathbf{I}$, we distill the teacher's reasoning into the student policy~$p_\theta(\cdot~\mid x)$ via OPSD on student-generated rollouts. However, standard distillation in this setting suffers from two degenerate copying behaviors. First, the teacher tends to reinforce, rather than correct, the student's self-repetitive loops~\citep{fang2026copy}. Second, the teacher often favors verbatim transcriptions of 
$\mathbf{I}$, which is intrinsically harmful since the student cannot access $\mathbf{I}$ at inference. To address this, we propose Repetition-Aware Distillation. We explicitly detect spans that either repeat earlier rollout content or transcribe $\mathbf{I}$. For these flagged tokens, we mask the standard distillation supervision and instead apply an unlikelihood penalty.

\textbf{Detecting Repetition and Transcription.}
We use token $n$-gram matching to detect both self-repetition and privileged-information transcription. Let $\mathcal{N}_n(\cdot)$ denote the set of token $n$-grams. We define privileged $n$-grams as those appearing in the $\mathbf{I}$ but not in the input $x$:
\begin{equation}
\mathcal{G}_{\mathrm{priv}}
= \mathcal{N}_n(x,\mathbf{I}) \setminus \mathcal{N}_n(x).
\label{eq:priv-ngrams}
\end{equation}
An $n$-gram in the student rollout $y$ is flagged if it belongs to $\mathcal{G}_{\mathrm{priv}}$ or has appeared earlier in $y$. To ignore accidental short overlaps, we filter these flagged tokens by length. The final binary mask $m_t$ is defined as:
\begin{equation}
m_t =
\begin{cases}
1, & \text{if token } t \text{ belongs to a contiguous flagged span of length } \ge L{\min}, \\
0, & \text{otherwise.}
\end{cases}
\label{eq:rep-mask}
\end{equation}
% The mask is held fixed during loss computation.
\textbf{Masked Distillation.} We use mask $m_t$ in Eq.~\eqref{eq:rep-mask} to exclude flagged tokens from OPSD in Eq.~\eqref{eq:opsd-trajectory}, replacing its uniform token average with an average over unmasked tokens: 
\begin{equation}
    \label{eq:rep-masked-trajectory}
    \ell_{\mathrm{distill}}(y)
    \triangleq
    \frac{1}{\max(1,N_y)}
    \sum_{t=1}^{|y|}(1-m_t)
    D_{\mathrm{KL}}\!\left(
        p_\theta(\cdot \mid x, y_{<t})
        \;\big\|\;
            p_\phi(\cdot \mid x, \mathbf{I}, y_{<t})
    \right),
\end{equation}
where $N_y \triangleq \sum_{t=1}^{|y|}(1-m_t)$ is the number
of unmasked tokens.
% The modified on-policy objective is
% \begin{equation}
%     \label{eq:rep-masked-loss}
%     \mathcal{L}_{\mathrm{distill}}
%     =
%     \mathbb{E}_{y \sim p_{\bar{\theta}}(\cdot \mid x)}
%     \left[\ell_{\mathrm{distill}}(y)\right],
%     \qquad
%     \bar{\theta} = \operatorname{sg}(\theta).
% \end{equation}
This changes only the token-level aggregation in OPSD:
rollout sampling and teacher stop-gradient remain unchanged.
Trajectories and masks are held fixed during optimization.
When $m_t=0$ for all $t$, the objective reduces to
Eq.~\eqref{eq:opsd}; fully masked trajectories contribute zero loss.

\textbf{Unlikelihood Penalty.} Masking removes distillation supervision at flagged positions, but does
not explicitly discourage the flagged tokens. We therefore apply an
unlikelihood penalty \citep{welleck2019neural}.
Let $p_t = p_\theta(y_t \mid x, y_{<t})$ be the student's probability
of the sampled token. The trajectory-level penalty is
\begin{equation}
    \label{eq:rep-unlikelihood-trajectory}
    \ell_{\mathrm{rep}}(y)
    \triangleq
    \frac{1}{\max(1,M_y)}
    \sum_{t=1}^{|y|} m_t \left[-\log\!\big(\max(1-p_t,\,\epsilon)\big)\right],
\end{equation}
Here, $\epsilon>0$ is a small floor ensuring numerical
stability. This clamping is essential: because degenerate repetition loops often occur with extremely high model confidence $p_t \to 1$, the raw term $-\log(1-p_t)$ and its gradients would otherwise diverge. 
% The on-policy unlikelihood objective is
% \begin{equation}
%     \label{eq:unlikelihood}
%     \mathcal{L}_{\mathrm{rep}}
%     =
%     \mathbb{E}_{y \sim p_{\bar{\theta}}(\cdot \mid x)}
%     \left[\ell_{\mathrm{rep}}(y)\right],
%     \qquad
%     \bar{\theta} = \operatorname{sg}(\theta).
% \end{equation}
% \begin{equation}
% \mathcal{L}
% =
% \mathcal{L}_{\mathrm{distill}}
% +
% \lambda_{\mathrm{rep}}\mathcal{L}_{\mathrm{rep}}.
% \label{eq:total-loss}
% \end{equation}
This term is zero when no flagged span is detected. The final objective is
\begin{equation}
    \label{eq:total-loss}
    \mathcal{L}_{\mathrm{rep-distill}}
    =
    \mathbb{E}_{y \sim p_{\bar{\theta}}(\cdot \mid x)}
    \left[
        \ell_{\mathrm{distill}}(y)
        +
        \lambda_{\mathrm{rep}}\,\ell_{\mathrm{rep}}(y)
    \right],
    \qquad
    \bar{\theta} = \operatorname{sg}(\theta).
\end{equation}
The two terms play complementary roles: masked distillation
preserves supervision outside detected repetitions and
transcriptions, while unlikelihood discourages the flagged
continuations.
Unless otherwise stated, we use $n=4$, $L_{\min}=32$,
$\lambda_{\mathrm{rep}}=0.1$, and $\epsilon=10^{-6}$.

\begin{algorithm}[t!]
\caption{SpatialOPSD}
\label{alg:spatial_opsd}
\begin{algorithmic}[1]
\Require Training set $\mathcal{D}$; model $p_\theta$; teacher $p_\phi$;
    spatial agent $\mathcal{A}$ with kernel $\mathcal{K}$
    and tools $\mathcal{T}$; privileged-information
    constructor $\Phi$ (Eq.~\eqref{eq:priv_ops})
\Require Training steps $U$; rollout limit $L_{\max}$;
    learning rate $\eta$; $n=4$; $L_{\min}=32$;
    $\lambda_{\mathrm{rep}}=0.1$; $\epsilon=10^{-6}$
\Ensure Trained model $p_\theta$

\Statex \textbf{Phase I: Verified Trace Summarization}
\State $\mathcal{D}_I \leftarrow \emptyset$
\For{each $(x,y^\star)\in\mathcal{D}$}
    \State $(\tau,y_{\mathcal{A}})\leftarrow
        \mathrm{RunAgent}(\mathcal{A},x;\mathcal{K},\mathcal{T})$
    \If{$\mathrm{Verify}(y_{\mathcal{A}},y^\star)$}
        \State $\mathcal{D}_I\leftarrow
            \mathcal{D}_I\cup\{(x,\Phi(\tau,y^\star))\}$
        \Comment{Construct privileged info from
            verified trace and label (Eq.~\eqref{eq:priv_ops})}
    \EndIf
\EndFor

\Statex \textbf{Phase II: Repetition-Aware Distillation}
\For{$u=1,\ldots,U$}
    \State Sample a minibatch $\mathcal{B}\subseteq\mathcal{D}_I$
    \State $\bar{\theta}\leftarrow\operatorname{sg}(\theta)$
    \For{each $(x,\mathbf{I})\in\mathcal{B}$}
        \State Sample $y\sim p_{\bar{\theta}}(\cdot\mid x)$
            until $\texttt{eos}$ or $L_{\max}$ tokens
        \State Build privileged $n$-gram set
            $\mathcal{G}_{\mathrm{priv}}$ from $(x,\mathbf{I})$
            using Eq.~\eqref{eq:priv-ngrams}
        \State Compute mask $m$ using
            Eq.~\eqref{eq:rep-mask}
        \Comment{Flags self-repetition and transcription of $\mathbf{I}$}
        \State Evaluate student $p_\theta$ and teacher $p_\phi$
            on prefixes $y_{<t}$
        \State Compute $\ell_{\mathrm{distill}}(y)$ and
            $\ell_{\mathrm{rep}}(y)$ using
            Eqs.~\eqref{eq:rep-masked-trajectory}
            and~\eqref{eq:rep-unlikelihood-trajectory}
    \EndFor
    \State Average trajectory losses over $\mathcal{B}$ to obtain
        $\widehat{\mathcal{L}}_{\mathrm{rep-distill}}$
    \State $\theta\leftarrow\theta-\eta\nabla_\theta
        \bigl(\widehat{\mathcal{L}}_{\mathrm{rep -distill}})$
    \Comment{Hold rollouts and masks fixed}
\EndFor
\State \Return $p_\theta$
\end{algorithmic}
\end{algorithm}

%% file: sections/5.Experiments.tex
\section{Experiments}
 % \vspace{-6pt}
\subsection{Training Data Construction and Baseline Setup} 
 % \vspace{-6pt}
% For a controlled comparison, we train SpatialOPSD, SFT, and GRPO on the same 5,008 MindCube instances, sharing identical training queries, student-visible inputs, and evaluation protocols, with method-specific configurations detailed separately. We strictly filter this subset to ensure every instance is solvable (verified by SpatialClaw-Qwen3.5-9B) and has a valid teacher rationale. SpatialOPSD's privileged records are also exclusively derived from this subset. For the SFT baseline, we use Qwen3.7-plus to generate tool-free rationales from standard inputs, retaining the shortest correct rationale from 8 samples per instance. For GRPO, we train using a sparse exact-match reward on the final \verb|\boxed{}| answer. Full formatting and training details are deferred to ~\ref{app:data}.

We construct the training sets from the 10,000-instance MindCube training split. For SpatialOPSD and GRPO, we use the same 5,008 instances on which SpatialClaw-Qwen3.5-9B produces a correct, ground-truth-verified answer. SpatialOPSD's privileged records are derived exclusively from the verified agent traces for these instances. For SFT, Qwen3.7-plus~\citep{qwen37plus} can not produce correct CoT solution for all of the same 5,008 instances. We therefore generate tool-free rationales with Qwen3.7-plus for the full 10,000-instance training split and retain 4,459 instances with a correct rationale for SFT. All three methods use the same student-visible input format and evaluation protocol. Full data construction, formatting, and training details are provided in~\ref{app:data}.

 % \vspace{-6pt}
\subsection{Evaluation Benchmarks}
 % \vspace{-6pt}
We evaluate SpatialOPSD on two complementary spatial reasoning benchmarks. MindCube~\citep{yin2025spatial} tests scene reconstruction and viewpoint simulation from partial observations, while ViewSpatial-Bench~\citep{li2026viewspatial} evaluates localization across egocentric and allocentric viewpoints. We report accuracy and the privileged-information leakage rate defined in Appendix~\ref{app:privileged_information_leakage}.

% \vspace{-20pt}
\subsection{Main Results}
% \vspace{-5pt}
Table~\ref{main_table_1} compares Spatial coding against proprietary models, open-source MLLMs, and tuning baselines on MindCube-Tiny and ViewSpatial. SpatialOPSD improves the Qwen3.5-9B baseline from 45.9 to 51.9, with specific efficacy in allocentric transformations (e.g., Among +11.7). By utilizing on-policy dense supervision from agent traces, it yields higher accuracy than standard SFT (47.6) and GRPO (48.2), effectively mitigating SFT's distribution shift and GRPO's sparse-reward limitations. Notably, in a single tool-free forward pass, SpatialOPSD retains 
91\% of the performance of its iterative, tool-dependent teacher, SpatialClaw (51.9 vs. 57.3). Despite its 9B size, the model achieves results competitive with larger proprietary systems—surpassing GPT-5~\citep{singh2025openai} and narrowing the gap to Gemini-3-Pro. These results suggest that internalizing complex agent trajectories provides a viable axis for capability enhancement alongside parameter scaling.

% \textbf{Improvements over the base model}. SpatialOPSD improves the Qwen3.5-9B baseline by +6.3 (45.6 to 51.9). The gains are most prominent in categories requiring allocentric layout and viewpoint transformations, such as \emph{Among} (+11.6) and \emph{Around} (+8.5).

% \textbf{Approaching the tool-dependent teacher}. The SpatialClaw agent achieves 57.3 via an iterative code-execution loop. In a single forward pass without tools, SpatialOPSD reaches 51.9, retaining approximately 90.6\% of the teacher's performance. This suggests that spatial reasoning capabilities can be transferred from an agent to a model through agent-generated supervision.

% \textbf{Outperforming SFT and RL baselines}. On the same 9B backbone, SpatialOPSD (51.9) surpasses standard SFT (47.6) and GRPO (48.2). We attribute this to our on-policy dense supervision from agent traces, which alleviates the distribution shifts of SFT and the sparse-reward issues of GRPO.

% \textbf{Competitiveness with proprietary models}. Despite its 9B backbone, SpatialOPSD performs favorably against larger proprietary models. It surpasses GPT-5 on average (51.9 vs. 51.0), with a +0.6 margin on MindCube-Tiny, and narrows the gap to Gemini-3-Pro (60.7). This demonstrates that internalizing agent traces is an effective strategy alongside parameter scaling.

\begin{table*}[t!]
\setlength{\tabcolsep}{4pt}
\centering
\caption{\small Comparison of the accuracy on spatial reasoning benchmarks. The best results are bolded. CP-RD: Camera perspective relative direction. CP-OVO: Camera perspective object view Orientation. PP-OVO: Person perspective object view orientation. PP-RD: Person perspective relative direction. PP-SS-RD: Person perspective scene simulation relative direction. }
\label{main_table_1}
\resizebox{\textwidth}{!}{%
\begin{tabular}{@{}l|c|cccc|cccccc@{}}
\noalign{\vspace{-1em}}
\toprule
  &\multicolumn{1}{c}{\textbf{Average}} & \multicolumn{4}{c}{\textbf{MindCube-Tiny}} & \multicolumn{6}{c}{\textbf{ViewSpatial}} \\
\hline
    & Avg. & Among & Around & Rotation & Overall & CP- & CP- & PP- & PP- & PP- & Overall \\ 
    &  &  & & & & RD & OVO & OVO & RD & SS-RD &  \\
   \hline
   \multicolumn{12}{c}{Proprietary Model} \\
   \hline
   GPT-5 & 51.0 & 38.2 & 68.4 & 94.5 & 56.3 & 60.2 & 27.9 & 41.0 & 48.5 & 40.1 & 45.6 \\
   Gemini-3-Pro & 60.7 & 60.7 & 77.2 & 93.0 & 70.9 & 61.9 & 31.6 & 41.1 & 74.3 & 38.9  & 50.4 \\
   \hline
   \multicolumn{12}{c}{Open-Source Model} \\
   \hline
   Qwen3-VL-8B-Instruct~\citep{bai2025qwen3} & 35.8 & 28.6 & 31.2 & 29.5 & 29.4 & 54.2 & 29.7 & 47.3 & 40.3 & 31.1 & 42.2 \\
   Cambrian-S-7B~\citep{yang2026cambrian} & 39.5 & 39.0 & 39.2 & 33.0 & 37.9 & 50.4 & 22.7 & 45.0 & 38.8 &  41.9 & 41.3 \\
   % InternVL3-8B & 40.1 & 40.3 & 63.2 & 37.0 & 41.5 & 50.3 & 27.5 & 42.6 & 37.5 & 27.3 & 38.7 \\
   InternVL3.5-8B~\citep{wang2025internvl3} & 40.1 & 38.2 & 44.4 & 34.5 & 40.2 & 49.8 & 24.7 & 50.3 & 34.6 & 32.9 & 40.0 \\
   VST-7B-SFT~\citep{yang2026visual} & 45.1 & 35.9 & 50.8 & 37.0 & 39.7 & 52.7 & 29.6 & 51.9 & 50.7 & 64.5 & 50.5\\
   % Qwen3.5-9B-CoT & 45.6 & 34.6 & 54.9 & 75.9 & 47.4 & 51.4 & 34.7 & 43.1 & 49.7 & 35.9 & 43.8 \\
   Qwen3.5-9B-CoT~\citep{qwen3.5} & 45.9 & 34.5 & 57.0 & 77.9 & 48.3 & 51.2 & 34.4 & 42.3 & 50.1 & 34.8 & 43.4 \\

    \hline
    \multicolumn{12}{c}{Spatial Coding Agent} \\
   \hline
    \rowcolor[HTML]{EAE8FD}SpatialClaw$_{\text{Qwen3.5-9B}}$~\citep{cho2026spatialclaw} & 57.3 & 57.7 & 76.4 & 61.5 & 62.9 & 60.3 & 36.7  & 50.7 & 63.5 & 42.6 & 51.6 \\
   \hline
    \multicolumn{12}{c}{Custom-Trained Model} \\
   \hline
    Qwen3.5-9B$_{\text{SFT}}$ & 47.6 & 50.4 & 60.6 & 77.7 & \textbf{58.1} & 44.3 & 30.7 & 38.1 & 37.8 & 29.3 & 37.0 \\
    Qwen3.5-9B$_{\text{GRPO}}$ & 48.2 & \textbf{51.2} & \textbf{68.0} & 41.2 & 53.3 & 54.3 & 31.1 & 46.1 & 47.6 & 29.6 & 43.1 \\
    \rowcolor[HTML]{EAE8FD} SpatialOPSD-9B (ours) & \textbf{51.9} &  46.2 & 63.4 & \textbf{80.5} & 56.9 & \textbf{56.4} & \textbf{35.7} & \textbf{48.9} & \textbf{50.6} & \textbf{36.6} & \textbf{46.8} \\
   \textcolor{gray}{\textit{$\Delta$ from Qwen3.5-9B-CoT}} & \small{\textcolor{blue}{+6.0}} & \small{\textcolor{blue}{+11.7}} & \small{\textcolor{blue}{+6.4}} & \small{\textcolor{blue}{+2.6}} & \small{\textcolor{blue}{+8.6}} & \small{\textcolor{blue}{+5.2}} & \small{\textcolor{blue}{+1.3}} &  \small{\textcolor{blue}{+6.6}} & \small{\textcolor{blue}{+0.5}} & \small{\textcolor{blue}{+1.8}} &  \small{\textcolor{blue}{+3.4}} \\
\bottomrule
\end{tabular}
}
\vspace{-1em}
\end{table*}

\subsection{Training Dynamics}

 \begin{figure*}[h]
        \centering
        \begin{subfigure}[t]{0.24\textwidth}
            \centering
            \includegraphics[width=\linewidth]{
                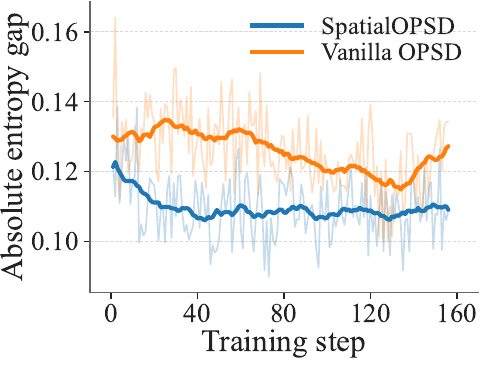
            }
            \caption{Entropy gap.}
            \label{fig:entropy_gap}
        \end{subfigure}
        \hfill
        \begin{subfigure}[t]{0.240\textwidth}
            \centering
            \includegraphics[width=\linewidth]{
                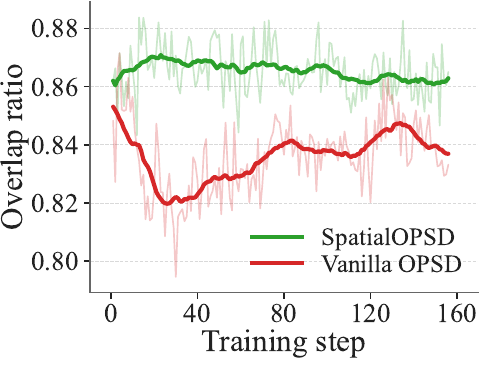
            }
            \caption{Overlap ratio. }
            \label{fig:overlap_ratio}
        \end{subfigure}
        \hfill
        \begin{subfigure}[t]{0.240\textwidth}
            \centering
            \includegraphics[width=\linewidth]{
                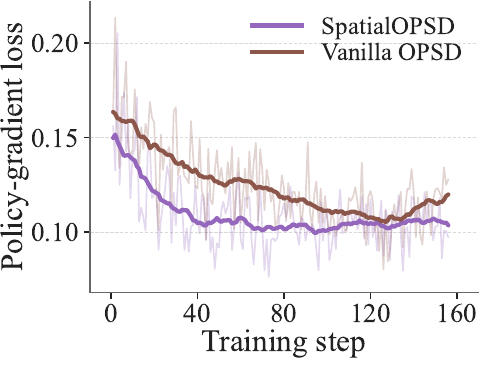
            }
            \caption{Policy-gradient Loss.}
            \label{fig:training_loss}
        \end{subfigure}
        \hfill
        \begin{subfigure}[t]{0.240\textwidth}
            \centering
            \includegraphics[width=\linewidth]{
                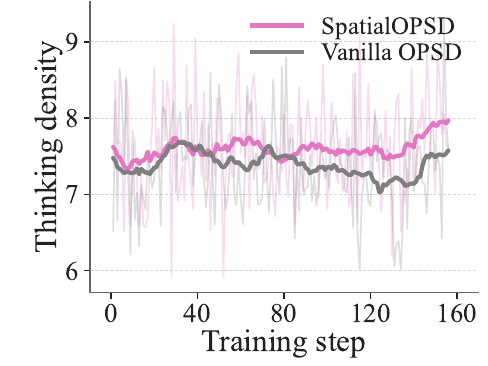
            }
            \caption{Thinking density.}
            \label{fig:thining_density}
        \end{subfigure}

        \caption{
            Training dynamics of SpatialOPSD-9B and vanilla OPSD, both using Qwen3.5-9B as the backbone.
        }
\label{fig:experiment_dynamics}
% \vspace{-10pt}
\end{figure*}

We report the training dynamics of SpatialOPSD and vanilla OPSD method using the original objective in Eq.~\eqref{eq:opsd} instead of Eq.~\eqref{eq:total-loss}, labeled “SpatialOPSD” and “Vanilla OPSD” in the Figure~\ref{fig:experiment_dynamics}, respectively. Figures~\ref{fig:entropy_gap} and~\ref{fig:overlap_ratio} show that SpatialOPSD achieves a smaller teacher--student entropy gap and greater overlap between their top-100 token sets, indicating closer distributional alignment. Figure~\ref{fig:training_loss} further shows faster and more stable loss convergence. Finally, Figure~\ref{fig:thining_density} reports a higher frequency of predefined spatial-reasoning terms in student rollouts, suggesting that SpatialOPSD more effectively elicits spatially relevant reasoning during training.

\subsection{Out-of-domain Evaluation}
% \vspace{-10pt}
% \begin{table*}[h!]
%     \centering
%     \small
%     \setlength{\tabcolsep}{12pt}
%     \renewcommand{\arraystretch}{1.15}

%     \caption{
%         Accuracy (\%) on four out-of-distribution (OOD) VQA benchmarks.
%         Average denotes the unweighted mean across benchmarks.
%         Bold indicates the best result in each row.
%     }
%     \label{tab:ood_vqa}

%     \begin{tabular}{@{}lccc>{\columncolor[HTML]{EAE8FD}}c@{}}
%         \toprule
%         & \multicolumn{4}{c}{\textbf{Qwen3.5-9B}} \\
%         \cmidrule(l){2-5}
%         \textbf{Benchmark}
%         & \textbf{CoT}
%         & \textbf{SFT}
%         & \textbf{GRPO}
%         & \textbf{SpatialOPSD} \\
%         \midrule

%         OmniSpatial
%         & 53.7 & 39.8 & 50.6 & \textbf{54.2} \\

%         BLINK
%         & 67.8 & 57.5 & 58.1 & \textbf{68.9} \\

%         MMStar
%         & \textbf{67.5} & 33.9 & 65.2 & 66.9 \\

%         VideoMME
%         & \textbf{68.7} & 32.3 & 62.4 & 68.6 \\

%         \midrule
%         \textbf{Average}
%         & 64.4 & 40.9 & 59.1 & \textbf{64.7} \\

%         \bottomrule
%     \end{tabular}
% \end{table*}

\begin{table*}[t]
    \centering
    \small
    \setlength{\tabcolsep}{6pt}
    \renewcommand{\arraystretch}{1.15}

\end{table*}

We evaluate SpatialOPSD on four held-out benchmarks unseen during training:
BLINK~\citep{fu2024blink} and OmniSpatial~\citep{jia2026omnispatial}
for spatial reasoning, and MMStar~\citep{chen2024we} and
Video-MME~\citep{fu2025video} for general multimodal capabilities.
As shown in Table~\ref{tab:ood_vqa}, SpatialOPSD achieves the
highest average accuracy across the four benchmarks (64.7\%),
outperforming both SFT and GRPO and slightly exceeding the
CoT baseline (64.4\%).
Its results on the spatial benchmarks, together with its
competitive performance on MMStar and Video-MME, suggest that
spatial reasoning gains do not substantially compromise performance
on the evaluated general multimodal benchmarks.

% \vspace{-4pt}

% \vspace{-6pt}
\subsection{Ablations}
% \vspace{-4pt}

% 正文
\newlength{\abltabheight}\setlength{\abltabheight}{35pt}
\begin{table*}[t]
    \centering

    % 左表：Privileged-input 与 distillation comparison
    \begin{minipage}[t]{0.4\textwidth}
        \centering
        \vspace{0pt}
        \setlength{\tabcolsep}{5pt}
        \caption{
        Accuracy (\%) on out-of-distribution (OOD) benchmarks.
        Bold indicates the best result in each row.
    }
    \label{tab:ood_vqa}
 \resizebox{!}{\abltabheight}{%
            \begin{tabular}{
                @{}lccc
                >{\columncolor[HTML]{EAE8FD}}c@{}
            }
                \toprule
                & \multicolumn{4}{c}{\textbf{Qwen3.5-9B}} \\
                \cmidrule(l){2-5}
                \textbf{Benchmark}
                & \textbf{CoT}
                & \textbf{SFT}
                & \textbf{GRPO}
                & \textbf{SpatialOPSD} \\
                \midrule

                OmniSpatial
                & 53.7 & 39.8 & 50.6 & \textbf{54.2} \\

                BLINK
                & 67.8 & 47.4 & 66.6 & \textbf{68.9} \\

                MMStar
                & 67.5 & 34.0 & \textbf{68.1} & 66.9 \\

                VideoMME
                & \textbf{68.7} & 32.3 & 68.4 & 68.6 \\

                \midrule
                \textbf{Average}
                & 64.4 & 38.4 & 63.4 & \textbf{64.7} \\
                \bottomrule
            \end{tabular}%
        }

    \end{minipage}
    \hfill
    % 右表：OPSD 组件消融
    \begin{minipage}[t]{0.55\textwidth}
        \centering
        \vspace{0pt}
        \setlength{\tabcolsep}{5pt}
        \caption{OPSD component ablation on masked distillation (Mask) and unlikelihood penalty (UL),
        reporting accuracy and privileged-information (PI) leakage.}
        \label{tab:opsd_ablation}

        \resizebox{!}{\abltabheight}{%
\begin{tabular}{@{}lcccccc@{}}
    \toprule
    \textbf{Method}
    & \textbf{Mask}
    & \textbf{UL}
    & \textbf{MindCube-}
    & \textbf{View}
    & \textbf{Avg.}
    & \textbf{PI} \\
    & & & \textbf{Tiny} & \textbf{Spatial} & (\%)
    & \textbf{leakage} \\
    & & & Overall (\%) & Overall (\%) & 
    & (\%) $\downarrow$ \\
    \midrule

    Qwen3.5-9B-CoT
    & -- & --
    & 48.3 & 43.4 & 45.9 & -- \\

    % \midrule
    % OPSD-PMI~\citep{shen2026purified}
    % & \xmark & \xmark
    % & 50.4 ({\small\textcolor{blue}{+3.0}})
    % & 44.9 ({\small\textcolor{blue}{+1.1}})
    % & 47.7 ({\small\textcolor{blue}{+2.1}})
    % & 20.0 \\

    % vigos~\citep{wang2026seeing}
    % & \xmark & \xmark
    % & 54.9 ({\small\textcolor{blue}{+7.5}})
    % & 43.6 ({\small\textcolor[HTML]{006400}{-0.2}})
    % & 49.3 ({\small\textcolor{blue}{+3.7}})
    % & 20.1 \\

    % dapd~\citep{wu2026dapd}
    % & \xmark & \xmark
    % & 54.9 ({\small\textcolor{blue}{+7.5}})
    % & 45.7 ({\small\textcolor{blue}{+1.9}})
    % & 50.3 ({\small\textcolor{blue}{+4.7}})
    % & 81.4 \\

    \midrule
    Vanilla OPSD
    & \xmark & \xmark
    & 55.4 ({\small\textcolor{blue}{+7.1}})
    & 46.0 ({\small\textcolor{blue}{+2.6}})
    & 50.7 ({\small\textcolor{blue}{+4.8}})
    & 82.7 \\

    Vanilla OPSD
    & \cmark & \xmark
    & 60.0 ({\small\textcolor{blue}{+11.7}})
    & 45.9 ({\small\textcolor{blue}{+2.5}})
    & 53.0 ({\small\textcolor{blue}{+7.1}})
    & 73.6 \\

    Vanilla OPSD
    & \xmark & \cmark
    & 55.6 ({\small\textcolor{blue}{+7.3}})
    & 47.3 ({\small\textcolor{blue}{+3.9}})
    & 51.5 ({\small\textcolor{blue}{+5.6}})
    & 6.7 \\

    \rowcolor[HTML]{EAE8FD}
    SpatialOPSD
    & \cmark & \cmark
    & 56.9 ({\small\textcolor{blue}{+8.6}})
    & 46.8 ({\small\textcolor{blue}{+3.4}})
    & 51.9 ({\small\textcolor{blue}{+6.0}})
    & 9.1 \\

    \bottomrule
\end{tabular}%
}

    \end{minipage}
\end{table*}
% \vspace{-5mm}

% \textbf{Impact of Privileged Information.}
% Table~\ref{tab:controlled_comparison} compares strategies for leveraging
% privileged information (PI).
% Providing \textsc{summary} as an additional training input degrades
% both SFT and GRPO, consistent with a train--test mismatch when PI
% is removed at inference.
% Off-policy distillation improves over the baseline, while SpatialOPSD
% further improves accuracy on both benchmarks.
% These results support distillation as an effective way to leverage PI
% without requiring it at inference.

\textbf{Impact of Repetition-Aware Distillation.} Table~\ref{tab:opsd_ablation} ablates masked distillation (Mask) and
unlikelihood regularization (UL).
Mask alone yields the highest average accuracy, while UL alone provides
greater leakage reduction.
Combining both substantially reduces PI leakage relative to vanilla OPSD, while trading some accuracy relative to Mask alone.
These results suggest complementary roles for Mask and UL in balancing
knowledge transfer and leakage suppression. The privileged information leakage rate is defined in~\ref{app:privileged_information_leakage}.

\setlength{\columnsep}{6pt}  % 水平间距，可进一步调小到 4pt
\setlength{\intextsep}{2pt}  % 上下间距
\begin{wraptable}[15]{r}{0.4\linewidth}
    \centering
        \caption{Effect of $L_{\min}$ and $\lambda_{rep}$ on accuracy and privileged information (PI) leakage rate. }
    \label{tab:robust}
    \resizebox{\linewidth}{!}{%
\begin{tabular}{@{}c|ccc@{}}
    \toprule
    \textbf{Value}
    & \textbf{MindCube-Tiny}
    & \textbf{ViewSpatial}
    & \textbf{PI Leakage} \\
    & Overall (\%) $\uparrow$
    & Overall (\%) $\uparrow$
    & (\%) $\downarrow$ \\
    \midrule

    \textbf{Baseline}
    & 55.4
    & 46.0
    & 82.7 \\

    \midrule
    \multicolumn{4}{c}{
        \textbf{Varying $\lambda_{\mathrm{rep}}$}
        \quad ($L_{\min}=32$)
    } \\
    \midrule

    0.00
    & 60.0 ({\small\textcolor{blue}{+4.6}})
    & 45.9 ({\small\textcolor[HTML]{006400}{-0.1}})
    & 73.6 ({\small\textcolor[HTML]{006400}{-9.1}}) \\

    0.05
    & 57.7 ({\small\textcolor{blue}{+2.3}})
    & 47.0 ({\small\textcolor{blue}{+1.0}})
    & 36.0 ({\small\textcolor[HTML]{006400}{-46.7}}) \\

    \rowcolor[HTML]{EAE8FD}
    0.10
    & 56.9 ({\small\textcolor{blue}{+1.5}})
    & 46.8 ({\small\textcolor{blue}{+0.8}})
    & 9.1 ({\small\textcolor[HTML]{006400}{-73.6}}) \\

    0.15
    & 55.1 ({\small\textcolor[HTML]{006400}{-0.3}})
    & 46.0 ({\small\textcolor{blue}{+0.0}})
    & 0.8 ({\small\textcolor[HTML]{006400}{-81.9}}) \\

    \midrule
    \multicolumn{4}{c}{
        \textbf{Varying $L_{\min}$}
        \quad ($\lambda_{\mathrm{rep}}=0.1$)
    } \\
    \midrule

    4
    & 56.0 ({\small\textcolor{blue}{+0.6}})
    & 46.9 ({\small\textcolor{blue}{+0.9}})
    & 75.9 ({\small\textcolor[HTML]{006400}{-6.8}}) \\

    8
    & 57.0 ({\small\textcolor{blue}{+1.6}})
    & 47.0 ({\small\textcolor{blue}{+1.0}})
    & 49.5 ({\small\textcolor[HTML]{006400}{-33.2}}) \\

    16
    & 56.4 ({\small\textcolor{blue}{+1.0}})
    & 47.0 ({\small\textcolor{blue}{+1.0}})
    & 24.0 ({\small\textcolor[HTML]{006400}{-58.7}}) \\

    \rowcolor[HTML]{EAE8FD}
    32
    & 56.9 ({\small\textcolor{blue}{+1.5}})
    & 46.8 ({\small\textcolor{blue}{+0.8}})
    & 9.1 ({\small\textcolor[HTML]{006400}{-73.6}}) \\

    64
    & 56.9 ({\small\textcolor{blue}{+1.5}})
    & 46.7 ({\small\textcolor{blue}{+0.7}})
    & 59.3 ({\small\textcolor[HTML]{006400}{-23.4}}) \\

    \bottomrule
\end{tabular}%
    }
    % \vspace{-2mm}

\end{wraptable}
\textbf{Impact of Hyperparameter.}
We ablate the repetition penalty weight $\lambda_{\mathrm{rep}}$ and
minimum repeated-span length $L_{\min}$ on MindCube-Tiny and ViewSpatial.
A moderate penalty suppresses PI leakage while improving accuracy;
a stronger penalty reduces leakage further at the expense of accuracy.
Varying $L_{\min}$ has little effect on accuracy but substantially
affects leakage.
We therefore use $\lambda_{\mathrm{rep}}=0.10$ and $L_{\min}=32$
to balance leakage suppression and accuracy
(Table~\ref{tab:robust}).

\textbf{Model Scale.}
We evaluate SpatialOPSD with Qwen3.5-2B, Qwen3.5-4B, and Qwen3.6-27B~\citep{qwen3.6-27b} students. Given the smaller models' limited coding-agent and tool-use capabilities, the 2B and 4B students train on data derived from verified SpatialClaw-Qwen3.5-9B traces, while the 27B student uses verified SpatialClaw-Qwen3.6-27B traces. As shown in Table~\ref{tab:general}, SpatialOPSD improves accuracy on both benchmarks at all three student scales. Average accuracy (the unweighted mean across benchmarks) increases from 40.6\% to 42.8\% at 2B, from 46.4\% to 50.2\% at 4B, and from 50.6\% to 54.2\% at 27B, corresponding to gains of 2.2, 3.8, and 3.6 percentage points, respectively.

% \vspace{-5mm}
% \textbf{Model Scale.} We evaluate SpatialOPSD on Qwen3.5 models at 2B, 4B, and 27B scales. As shown in Table~\ref{tab:general}, SpatialOPSD consistently improves accuracy on both benchmarks across all three scales. Average accuracy (the unweighted mean across benchmarks) increases from 40.6\% to 42.8\% at 2B, from 46.4\% to 50.2\% at 4B, and from 50.6\% to 54.2\% at 27B, corresponding to gains of 2.2, 3.8, and 3.6 percentage points, respectively.

\begin{table}[h!]
\centering
% Sequence length ablation
% 左表
\begin{minipage}[t]{0.48\linewidth}
    \centering
        \caption{Performance Comparison of SpatialOPSD across different model sizes.}
    \label{tab:general}
    % \vspace{0pt}
    \resizebox{\linewidth}{!}{%
 \begin{tabular}{@{}l|ccc@{}}
    \toprule
    \textbf{Model}
    & \textbf{MindCube-Tiny}
    & \textbf{ViewSpatial}
    & \textbf{Avg.} \\
    & \textbf{Overall} (\%)
    & \textbf{Overall} (\%)
    & (\%) \\
    \midrule
    Qwen3.5-2B
    & 41.4 & 39.8 & 40.6 \\
    \rowcolor[HTML]{EAE8FD} SpatialOPSD-2B
    & \textbf{43.5} ({\small\textcolor{blue}{+2.1}})
    & \textbf{42.1} ({\small\textcolor{blue}{+2.3}})
    & \textbf{42.8} ({\small\textcolor{blue}{+2.2}}) \\
    \midrule
    Qwen3.5-4B
    & 47.7 & 45.1 & 46.4 \\
    \rowcolor[HTML]{EAE8FD} SpatialOPSD-4B
    & \textbf{53.8} ({\small\textcolor{blue}{+6.1}})
    & \textbf{46.6} ({\small\textcolor{blue}{+1.5}})
    & \textbf{50.2} ({\small\textcolor{blue}{+3.8}}) \\
    \midrule
    Qwen3.6-27B
    & 54.1 & 47.1 & 50.6 \\
    \rowcolor[HTML]{EAE8FD} SpatialOPSD-27B
    & \textbf{58.8} ({\small\textcolor{blue}{+4.7}})
    & \textbf{49.6} ({\small\textcolor{blue}{+2.5}})
    & \textbf{54.2} ({\small\textcolor{blue}{+3.6}}) \\
    \bottomrule
\end{tabular}%
    }

\end{minipage}%
\hfill
% Privileged information type ablation
\begin{minipage}[t]{0.48\linewidth}
    \centering
    
    \setlength{\tabcolsep}{6pt}
        \caption{Effect of privileged information types.}
    \label{tab:privileged_information_types}
    \vspace{12pt}

    \resizebox{\linewidth}{!}{%
    \begin{tabular}{@{}lccc@{}}
    \toprule
    \textbf{Method / PI Type}
    & \textbf{MindCube-Tiny}
    & \textbf{ViewSpatial}
    & \textbf{Avg.} \\
    & Overall (\%) & Overall (\%) & (\%) \\
    \midrule

    Qwen3.5-9B-CoT
    & 48.3 & 43.4 & 45.9 \\

    \midrule
    \multicolumn{4}{@{}l}{
        \textit{SpatialOPSD with different PI types}
    } \\[2pt]

    \quad \textsc{intent}
    & 53.3 ({\small\textcolor{blue}{+5.0}}) & \textbf{47.4} ({\small\textcolor{blue}{+4.0}}) & 50.4 ({\small\textcolor{blue}{+4.5}}) \\

    \quad \textsc{full-trace}
    & 54.5 ({\small\textcolor{blue}{+6.2}}) & 45.8 ({\small\textcolor{blue}{+2.4}}) & 50.2 ({\small\textcolor{blue}{+4.3}}) \\

    \quad \textsc{fact}
    & 55.2 ({\small\textcolor{blue}{+6.9}}) & 45.4 ({\small\textcolor{blue}{+2.0}}) & 50.3 ({\small\textcolor{blue}{+4.4}}) \\

    \rowcolor[HTML]{EAE8FD}
    \quad \textsc{summary}
    & \textbf{56.9} ({\small\textcolor{blue}{+8.6}}) & 46.8 ({\small\textcolor{blue}{+3.4}}) & \textbf{51.9} ({\small\textcolor{blue}{+6.0}}) \\

    \bottomrule
\end{tabular}%
}

    % \caption{Effect of privileged information types.}
    % \label{tab:privileged_information_types}
\end{minipage}
\end{table}

\begin{figure*}[t!]
        \centering
        \begin{subfigure}[t]{0.32\textwidth}
            \centering
            \includegraphics[width=\linewidth]{
                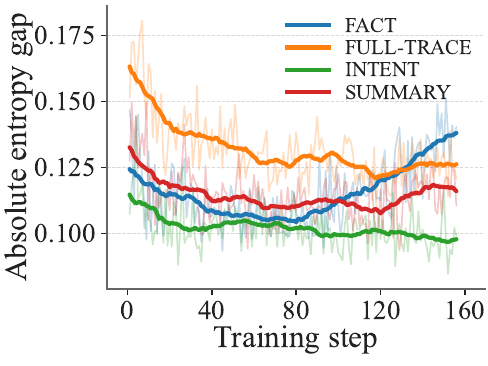
            }
            \caption{Entropy difference.}
            \label{fig:experiment_down}
        \end{subfigure}
        \hfill
        \begin{subfigure}[t]{0.32\textwidth}
            \centering
            \includegraphics[width=\linewidth]{
                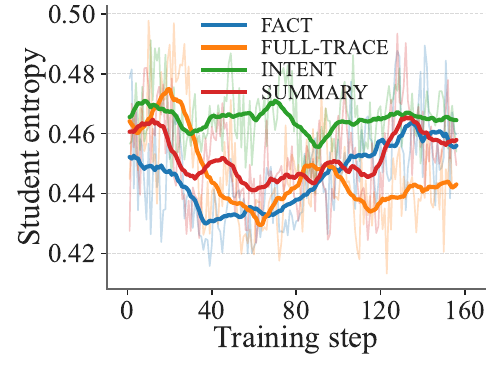
            }
            \caption{Student entropy.}
            \label{fig:experiment_down}
        \end{subfigure}
        \hfill
        \begin{subfigure}[t]{0.32\textwidth}
            \centering
            \includegraphics[width=\linewidth]{
                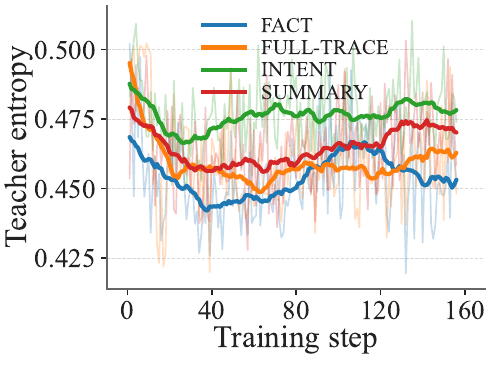
            }
            \caption{Teacher entropy.}
            \label{fig:experiment_down}
        \end{subfigure}

        \caption{
           Training dynamics under different PI variants. We compare FULL-TRACE, SUMMARY, FACT, and INTENT in terms of (a) absolute entropy gap, (b) student entropy, and (c) teacher entropy.
        }
        \label{fig:privileged information type}
\end{figure*}

% \clearpage
\textbf{Privileged Information Types.} We also evaluate how different strategies for constructing privileged information affect SpatialOPSD. The results in Table~\ref{tab:privileged_information_types} show that \textsc{summary} achieves the best performance. Specifically, in Figure~\ref{fig:privileged information type}, we find that \textsc{summary} maintains a moderate entropy gap between the student and teacher. Although \textsc{intent} yields a smaller gap, both models exhibit high entropy. In contrast, \textsc{full-trace} induces a large entropy gap that hinders optimization, while \textsc{fact} exhibits a growing gap in the later stages of training, likewise compromising optimization outcomes.

% \textbf{Forward vs. reverse KL divergence.}
% Impact of the KL-divergence direction on the performance of SpatialOPSD on MindCube. As shown in Table~\ref{tab:kl_divergence}, reverse KL achieves better performance than forward KL.

% \begin{tabular}{@{}lc@{}}
% \toprule
% \textbf{Method} &  MindCube Overall  \\
% \midrule
% \textbf{Forward KL} & 54.7 \\
% \textbf{Reverse KL} & 57.8 \\
% \bottomrule
% \end{tabular}

\textbf{Effect of verification.}
As shown in Table~\ref{tab:verification}, training without verification uses agent traces from all 10,000 MindCube training samples, whereas verification retains only the 5,008 traces whose final answers match the ground truth. Despite using fewer traces, the verified subset improves performance by 4.9 points on MindCube-Tiny and 0.5 points on ViewSpatial, highlighting the importance of trace quality over quantity.

\textbf{Scalability across coding agents.}
Table~\ref{tab:coding_agent} shows that SpatialOPSD generalizes across different coding agents. Traces generated by both SpatialClaw~\citep{cho2026spatialclaw} and pySpatial~\citep{luo2026pyspatial}  consistently improve the base model, while pySpatial achieves comparable or slightly better performance, demonstrating the agent-agnostic scalability of our framework.

% \begin{table}[h!]
% \centering

% \setlength{\tabcolsep}{4pt}
% \resizebox{\textwidth}{!}{%
% \begin{tabular}{@{}cc|cccc|cccccc@{}}
% \toprule
% & & \multicolumn{4}{c}{\textbf{MindCube-Tiny}} & \multicolumn{5}{c}{\textbf{ViewSpatial}}\\
% Model & \textbf{Verification}
% & \textbf{Among}
% & \textbf{Around}
% & \textbf{Rotation}
% & \textbf{Overall} & \textbf{CP-RD} & \textbf{CP-OVO} & \textbf{PP-OVO} & \textbf{PP-RD} & \textbf{PP-SS-RD} & \textbf{Overall} \\
% \midrule
% % \xmark & 30.95 & 45.04 & 69.30 & 41.71 \\
% % \cmark & 48.68 & 52.88 & 49.50 & 49.85 \\
% Qwen3.5-9B-CoT & - & 34.5 & 57.0 & 77.9 & 48.3 & 51.2 & 34.4 & 42.3 & 50.1 & 34.8 & 43.4 \\
% SpatialOPSD-9B & \xmark & 42.1 & 59.9 & 71.4 & 52.0 & \textbf{56.4} & 35.4 & \textbf{48.9} & 48.2 & 35.9 & 46.3 \\
% % SpatialOPSD-9B & \cmark & 47.8 & 65.8 & 77.2 & 57.8 & 56.0 & {35.3} & {48.5} & 50.8 & 33.8 & {46.0} \\
%     \rowcolor[HTML]{EAE8FD} SpatialOPSD-9B & \cmark & \textbf{46.2} & \textbf{63.4} & \textbf{80.5} & \textbf{56.9} & \textbf{56.4} & \textbf{35.7} & \textbf{48.9} & \textbf{50.6} & \textbf{36.6} & \textbf{46.8} \\

% \bottomrule
% \end{tabular}
% }
% \caption{Effect of verification using agent execution results and ground-truth answers.}
% \label{tab:verification}
% \end{table}
\begin{table*}[h]
\centering

\begin{minipage}[t]{0.49\textwidth}
\centering
\setlength{\tabcolsep}{4pt}
\caption{Effect of verification using agent execution results and ground-truth answers.}
\label{tab:verification}
\resizebox{\linewidth}{!}{%
\begin{tabular}{@{}cc|cc|c@{}}
\toprule
& & \textbf{MindCube-Tiny} & \textbf{ViewSpatial} & \\
Model & \textbf{Verification}
& \textbf{Overall} & \textbf{Overall} & \textbf{Avg} \\
\midrule
Qwen3.5-9B-CoT & - & 48.3 & 43.4 & 45.9 \\
SpatialOPSD-9B & \xmark & 52.0 & 46.3 & 49.2 \\
\rowcolor[HTML]{EAE8FD}
SpatialOPSD-9B & \cmark
& \textbf{56.9} & \textbf{46.8} & \textbf{51.9} \\
\bottomrule
\end{tabular}%
}
% \caption{Effect of verification using agent execution results and ground-truth answers.}
% \label{tab:verification}
\end{minipage}
\hfill
\begin{minipage}[t]{0.49\textwidth}
\centering
\setlength{\tabcolsep}{4pt}
\caption{Performance of SpatialOPSD trained on different coding agent traces.}
\label{tab:coding_agent}
\resizebox{\linewidth}{!}{%
\begin{tabular}{@{}cc|cc|c@{}}
\toprule
& & \textbf{MindCube-Tiny} & \textbf{ViewSpatial} & \\
Model & \textbf{Coding Agent}
& \textbf{Overall} & \textbf{Overall} & \textbf{Avg} \\
\midrule
Qwen3.5-9B-CoT & - & 48.3 & 43.4 & 45.9 \\
\rowcolor[HTML]{EAE8FD}SpatialOPSD-9B & SpatialClaw
& 56.9 & 46.8 & 51.9 \\

SpatialOPSD-9B & pyspatial
& \textbf{57.0} & \textbf{47.1} & \textbf{52.0} \\
\bottomrule
\end{tabular}%
}
% \caption{Performance of SpatialOPSD trained on different coding agent traces.}
% \label{tab:coding_agent}
\end{minipage}

\end{table*}

%% file: sections/6.Analysis.tex
% \vspace{-5mm}
\section{Analysis}
% \vspace{-3mm}
\textbf{Quantitative Analysis.} To examine how training changes outputs, we analyze reasoning traces on 1,040 MindCube-tiny samples. Evaluated by lexical marker frequency and rollout coverage, training consistently increases all eight target categories. The model adopts a more structured reasoning style: Step Numbering exhibits the largest coverage gain (+53.0 points), followed by Verification (+43.9) and Scene Context (+33.8). While these metrics capture organizational changes in the generated traces, they measure articulation style rather than intrinsic reasoning ability (details in \ref{app:analysis}).

\vspace{+3mm}

\begin{figure}[!h]
    \centering
    \includegraphics[width=1.0\linewidth]{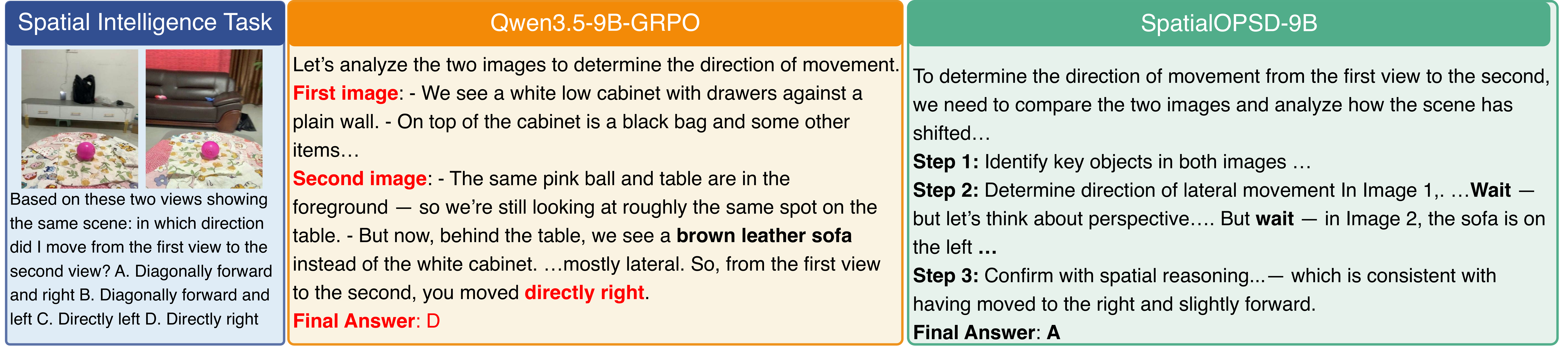}
    \vspace{-5mm}
    \caption{Performance comparison of Qwen3.5-9B-GRPO and SpatialOPSD-9B on MindCube-Tiny.}
    \label{fig:spatialopsd_case}
    % \vspace{-3mm}
\end{figure}

% \subsection{What does OPSD Learn} 

\begin{figure}[!h]
    \centering
    \includegraphics[width=1.0\linewidth]{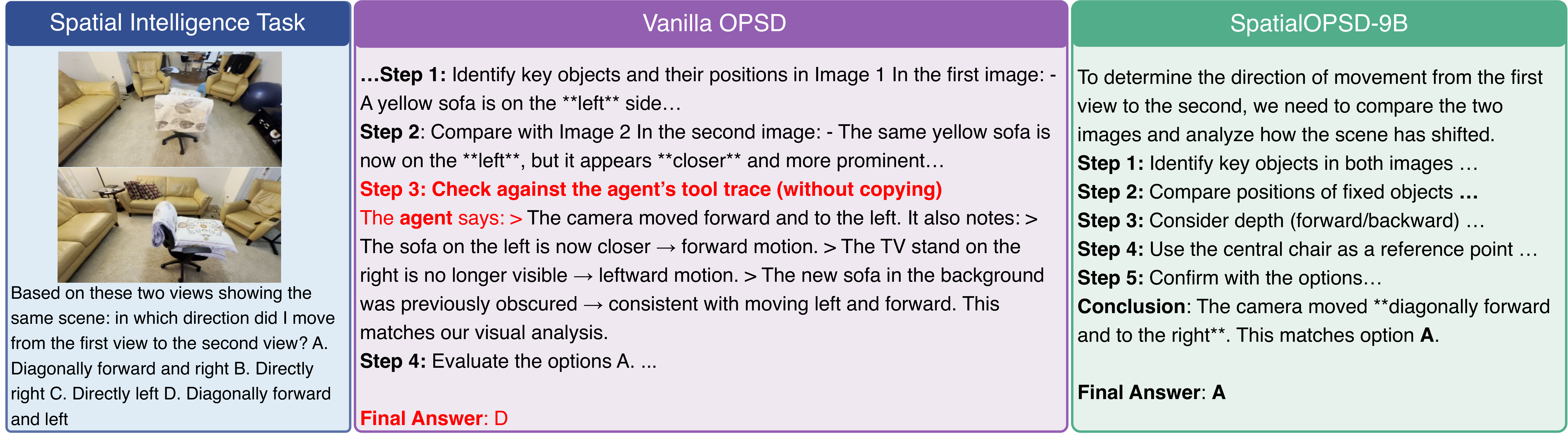}
    % \vspace{-5mm}
    \caption{\textbf{Privileged information leakage.}
The baseline generates unexecutable agent-like traces, whereas
SpatialOPSD-9B produces grounded spatial reasoning.}
    \label{fig:spatialopsd_failure_case}
    % \vspace{-3mm}
\end{figure}

% \vspace{+3mm}

\textbf{Qualitative Analysis.}
Figures~\ref{fig:spatialopsd_case} and~\ref{fig:spatialopsd_failure_case} illustrate the behavioral advantages of SpatialOPSD-9B. First, unlike Qwen3.5-9B-GRPO which conflates viewpoint shifts with simple 1D motion (purely rightward), our model correctly disentangles multi-axis trajectories. Second, vanilla OPSD suffers from shortcut learning, hallucinating unavailable tool traces to justify errors. In contrast, SpatialOPSD grounds its inference strictly in observable visual cues, such as depth and relative positions. These cases demonstrate that our method enables true viewpoint-aware reasoning while effectively mitigating spurious reliance on privileged training artifacts.

%% file: sections/2.Related.tex
\section{Related Work}
% \vspace{-4mm}
% \textbf{Spatial Intelligence.} To infer geometric structures and object relations, recent MLLMs typically rely on executable spatial tools. Methods like S-Agent~\citep{dai2026s} demonstrate the value of such interactions, yet they require the agent to actively invoke tools during deployment. Unlike these approaches, SpatialOPSD strictly confines tool usage to the training phase. Specifically, verified execution traces are summarized as privileged information for the teacher, while the student observes only the original multimodal input. By design, our goal is not to train another tool-using agent, but to seamlessly transfer tool-derived supervision to a standalone, tool-free model.

% 空间理解：
% 大规模qa训练模型：
% spatialvlm：https://arxiv.org/abs/2401.12168
% MM-Spatial：https://arxiv.org/abs/2503.13111
% SenseNova-SI: https://arxiv.org/abs/2511.13719
% Holi-Spatial：https://arxiv.org/abs/2603.07660
% OpenSpatial: https://arxiv.org/abs/2604.07296
% SpatialScore:  https://arxiv.org/abs/2505.17012

% 先sft再强化：
% ProSR：https://arxiv.org/abs/2605.25524
% mindcube：https://arxiv.org/abs/2506.21458
% SpaceVista：https://arxiv.org/abs/2510.09606
% ViGoRL：https://arxiv.org/abs/2505.23678

% training free的空间理解Agent：
% Think3D：https://arxiv.org/abs/2601.13029
% pySpatial:    https://arxiv.org/abs/2603.00905
% spatialclaw：   https://arxiv.org/abs/2606.13673
% GCA：   https://arxiv.org/pdf/2511.22659
% ViSRA：    https://arxiv.org/pdf/2605.10106
% SpatialScore：    https://arxiv.org/abs/2505.17012
% S-agent： https://arxiv.org/abs/2606.20515

\textbf{Spatial Intelligence in Multimodal LLMs.}
Recent spatial MLLMs generally follow three paradigms. (i) Supervised fine-tuning on QA pairs~\citep{chen2024spatialvlm, daxberger2025mm, cai2026scaling, gao2026holi, liu2026openspatial, wu2026spatialscore} generalizes poorly beyond training data and leaves spatial reasoning implicit. (ii) RL-optimized Chain-of-Thought~\citep{yin2025spatial, sarch2026grounded, sun2025spacevista, li2026prosr} provides explicit logic, but involves a complex pipeline bottlenecked by the high cost of curating cold-start trajectories.  (iii) Tool augmentation~\citep{mou2026visra, dai2026s, zhang2026think3d, luo2026pyspatial, cho2026spatialclaw} avoids training, yet suffers from heavy inference overhead and compounding cascading errors. In this paper, we propose a self-distillation approach that leverages agent traces to elicit and internalize spatial CoT reasoning.

\textbf{On-Policy Self-Distillation.} On-policy self-distillation (OPSD) trains a student via a teacher conditioned on privileged information(e.g., reference solutions~\citep{zhao2026self}, environment feedback~\citep{hubotter2026reinforcement}, or—in multimodal extensions—high-resolution visual crops \citep{yuan2026vision, cai2026thinking, zhu2026rp} and reasoning rationales \citep{li2026visual, cheng2026omniopsd}). A fundamental bottleneck is privilege leakage \citep{yang2026self}:  the student hallucinates unobserved references at inference~\citep{wu2026dapd}). To mitigate this, prior solutions typically modify the teacher's supervision by making the privilege input-recoverable \citep{tian2026vicur, zhao2026more}, temporally localized \citep{wang2026seeing}, or analytically isolated \citep{wu2026dapd}. In SpatialOPSD, rather than redesigning the teacher's supervision, we leave the privilege intact and simply penalize student tokens that reproduce trace-exclusive content. 

\vspace{-3mm}

%% file: sections/7.Conclusion.tex
% \vspace{-3mm}

\section{Conclusion}
% \vspace{-3mm}
We introduced SpatialOPSD, which distills verified spatial coding agent traces into MLLMs as privileged information for on-policy self-distillation. To separate spatial reasoning from tool-specific artifacts, Repetition-Aware Distillation penalizes copying privileged content and self-repetition. On MindCube and ViewSpatial, SpatialOPSD outperforms SFT and RL baselines and approaches tool-using agents without inference-time tools. By using training data generated by the same MLLM, SpatialOPSD forms a self-distillation learning loop, offering a practical route from tool-assisted data generation to stronger standalone spatial reasoning.

%% file: appendix/appendix.tex
\section{Appendix}

% \subsection{LLM Usage}
% Large Language Models (LLMs) were used solely for linguistic assistance in manuscript preparation, including sentence rephrasing, grammar checking, and improving readability. The LLM had no involvement in research ideation, methodology, experimental design, or data analysis. All scientific content and analyses were developed by the authors, who take full responsibility for the manuscript content and ensure compliance with ethical guidelines.

\subsection{Construction Details of Privileged Information}
\label{app:priv-construction}

\paragraph{Overview.}
Privileged information is harvested from successful rollouts of \emph{SpatialClaw}, a tool-using spatial-reasoning agent whose backbone is the same Qwen3.5-9B model that we later train. The pipeline has four stages.

\emph{Stage 1 (rollout and filtering).} We run SpatialClaw on the MindCube training split ($10{,}000$ multi-view spatial questions; settings \texttt{among}/\texttt{around}/\texttt{rotation} $=8{,}056/1{,}080/864$). For reference we also run the same backbone as a tool-free chain-of-thought (CoT) baseline on the identical split under identical decoding settings. The agent answers $5{,}008$ questions correctly ($50.08\%$) versus $4{,}384$ ($43.84\%$) for the tool-free baseline, and is correct on $2{,}088$ questions that the tool-free baseline gets wrong. We retain the $5{,}008$ agent-correct sessions; every privileged variant in this paper is built from exactly this set, so all variants are matched sample-for-sample.

\emph{Stage 2 (raw privileged context).} Each retained session log is parsed into a \emph{raw privileged context}: the original question and input frames, the agent's execution plan, every tool-using step (purpose, reasoning, next goal, code, and execution output), every \texttt{show()} visualization the agent actually viewed, and a final-answer section carrying both the agent's answer and the ground-truth label. The per-session Agent System Prompt (tool API documentation) is byte-identical across samples and carries no instance-specific signal, so it is removed at this stage.

\emph{Stage 3 (variant construction).} From this shared raw context we derive four variants of privileged information, summarized in Table~\ref{tab:priv-variants}. Full-trace is assembled deterministically and calls no LLM. Summary, Fact, and Intent are produced by prompting \textbf{Qwen3.5-9B} once per sample as an annotator, followed by deterministic post-processing:
\begin{itemize}
  \item \textbf{Answer scrubbing} (Summary, Fact, Intent). A line-level filter removes any residual answer declaration (a line starting with ``final answer'', ``answer:'', or ``the answer'') that the annotator emitted despite the no-leak instruction.
  \item \textbf{Citation tokenization} (Summary, Fact). Every file path the annotator cited verbatim is assigned a stable index in order of first citation. Its first occurrence is rewritten to a real image token \texttt{<image> (image $N$)}, and every later reference to the same path becomes the handle \texttt{(image $N$)}. The cited paths, in citation order, become that sample's privileged image list. Because an annotator may cite an \emph{original} input frame in order to redirect attention to it, the teacher can receive a frame twice: once at full training resolution at the top of the prompt, and once inline as a cited thumbnail. A separate pass downscales any re-cited full-resolution original to a $400$\,px thumbnail so that re-citation costs few tokens.
  \item \textbf{Strategy sanitization} (Intent). Any accidentally cited original-frame path is rewritten to ``image~$k$'', and any line referencing an agent visualization is dropped in whole. The Intent privileged body therefore carries no image token at all, and the teacher sees exactly the student's frames.
  \item \textbf{Final-answer append} (Fact). The trace's final-answer section is appended back onto the distilled body, byte-identically to the Full-trace formatting. This is a deterministic post-process rather than a relaxation of the annotation prompt, which keeps the distilled body identical to the audited version and makes the answer section comparable across Fact and Full-trace.
\end{itemize}

\emph{Stage 4 (OPSD record packing).} Each variant is packed into On-Policy Privileged Distillation (OPSD) training records with a shared student prompt and a variant-specific teacher prompt:
\begin{itemize}
  \item \textbf{Student prompt} (identical across all four variants): one \texttt{<image>} token per original frame, then the question and options with all dataset-specific direct-answer instructions stripped, then the chain-of-thought suffix ``\texttt{Please reason step by step, and put your final answer (the option's letter) within \textbackslash boxed\{\}.}''. Stripping the MindCube direct-answer rule is required and not merely cosmetic: its worked example quotes option letters (\texttt{'A.\ above'} from \texttt{'A.\ above B.\ under C.\ front D.\ behind.'}) that would otherwise contaminate the option list.
  \item \textbf{Teacher prompt}: the same question core, then a variant-specific wrapper that introduces the privileged block, then the same chain-of-thought suffix. The teacher additionally receives whatever privileged images the variant retains.
\end{itemize}
Original frames are resized to a maximum long edge of $2048$\,px, matching the resolution protocol used by all of our training and evaluation runs. Every record is checked against three invariants before it is written: the number of \texttt{<image>} tokens equals the number of images on both the student and the teacher side, the question core is a prefix of the teacher prompt, and the ground-truth label is non-empty.

At training time, a flat prompt string is split on \texttt{<image>} and interleaved with the corresponding images into a single user turn, which the Qwen3.5 chat template renders as \texttt{<|im\_start|>user}\,\ldots\,\texttt{<|im\_end|>} followed by the assistant generation prompt; under our no-thinking configuration the template additionally emits an empty \texttt{<think></think>} block. Each \texttt{<image>} becomes \texttt{<|vision\_start|><|image\_pad|><|vision\_end|>}, which the processor expands into image-patch tokens. The listings below report the user-message content that is inserted into this wrapper, which is the only part that differs across variants.

\begin{table}[t]
    \centering
    \small
    \caption{The four privileged-information variants, all constructed from the same $5{,}008$ agent-correct SpatialClaw traces. ``Teacher images'' counts what the teacher receives \emph{in addition} to the original input frames that the student also sees.}
    \label{tab:priv-variants}
    \setlength{\tabcolsep}{3pt}

    \begin{tabular}{lllcl}
        \toprule
        Variant & Constructor & Privileged body & Answer stated & Teacher images \\
        \midrule
        Full-trace & Deterministic & Raw execution log & Yes & Agent \texttt{show()} frames \\
        Summary    & Qwen3.5-9B, \texttt{style=full}   & Key reasoning points   & No  & Cited images \\
        Fact       & Qwen3.5-9B, \texttt{style=facts}  & Tool-measured facts    & Appended & Cited images \\
        Intent     & Qwen3.5-9B, \texttt{style=intent} & Tool-free strategy     & No  & None \\
        \bottomrule
    \end{tabular}
\end{table}

The remainder of this section gives, for each variant, the complete construction prompt (or the deterministic procedure, for Full-trace) followed by one fully instantiated OPSD example. All four examples use the same MindCube training instance, \texttt{among\_group542\_q3\_1\_1}: a two-view camera-motion question whose SpatialClaw trace calls a 3D reconstruction tool, reads out metric camera poses, and renders a bird's-eye-view (BEV) ego trajectory. Because the student prompt is variant-independent, Listings~\ref{lst:ft-student}, \ref{lst:sum-student}, \ref{lst:fact-student} and \ref{lst:int-student} are identical; only the teacher-side privileged block changes.

\begin{lstlisting}[caption={Qwen3.5-9B user message, shared by the Summary and Fact variants. Placeholders in braces are filled per sample. Because every retained trace is agent-correct, the ``Agent outcome'' field is always CORRECT on this dataset and the wrong-answer branch of the system prompts is never exercised.},label={lst:sum-user}]
## Question
{question}

## Ground-truth answer
{ground_truth_answer}

## Agent outcome
The agent's final answer was CORRECT.

## Original input frames (cite any of these by their EXACT path to focus attention on that frame)
  [1] {absolute_path_of_original_frame_1}
  [2] {absolute_path_of_original_frame_2}
  ...

## Privileged agent reasoning trace
(Visualization images appear inline as <<VIS IMAGE file=...>>.)

{raw privileged context, with the k-th <image> token replaced by
 <<VIS IMAGE file={show_images[k]}>>}
\end{lstlisting}

\subsubsection{Version of Full-trace}

The Full-trace variant does \emph{not} query Qwen3.5-9B. Starting from the raw privileged context we delete only the byte-identical Agent System Prompt span and keep everything else verbatim: the execution plan (task analysis, information needs, computation plan, verification checklist, fallbacks), every step including failed ones, every \texttt{show()} visualization in the order the agent viewed it, and the final-answer section. This variant is therefore the upper bound on privileged content: it is the only variant whose body retains code, tool-call mechanics, exploratory detours, and the ground-truth label as produced by the agent.

\begin{lstlisting}[caption={Deterministic construction of the Full-trace privileged body. No LLM is involved.},label={lst:ft-construct}]
Input : raw privileged context C parsed from an agent-correct SpatialClaw session
Output: Full-trace privileged body P

1. Locate the span "\n\n## Agent System Prompt\n" + safe(system_prompt) where system_prompt is the session's init-event system prompt and safe(.) rewrites literal "<image>" / "<video>" tokens inside it to "[image]" / "[video]" so they are never counted as vision placeholders.
2. Delete that span. Everything else in C is kept verbatim:
     - the header "Privileged Agent Reasoning Trace (SpatialClaw)"
     - "## Execution Plan"
     - "## Step-by-step Execution" (every step, including failed ones)
     - every show() visualization, in the order the agent viewed it
     - "## Final Answer" (agent final answer + ground-truth answer)
3. P <- the remaining text. Teacher images <- original frames ++ show() frames.
\end{lstlisting}

\begin{lstlisting}[caption={Teacher-side wrapper for the tool-trace variants (Full-trace, Summary, Fact). Reproduced verbatim.},label={lst:trace-wrapper}]
{question core, i.e. <image> tokens + question + options}

Here is an execution trace from a vision-tool-augmented agent that analyzed the same input images with perception tools:
=== Agent Tool-Trace Begin ===
{privileged body}
=== Agent Tool-Trace End ===

After reading the agent's tool-trace above, make sure you truly understand what each measured fact looks like in the images --- do not copy or cite the trace itself. Now, verify the facts against the original images with your own eyes and, using your own words and independent reasoning, derive the answer to the question above. Think step by step, explore different interpretations, and don't be afraid to backtrack or re-examine the images if something doesn't work out.

Please reason step by step, and put your final answer (the option's letter) within \boxed{}.
\end{lstlisting}

\paragraph{OPSD example.} Instance \texttt{among\_group542\_q3\_1\_1}; ground truth \texttt{A}. The student receives $2$ images, the teacher $3$.

\begin{lstlisting}[caption={Full-trace student prompt. Identical across all four variants.},label={lst:ft-student}]
<image>
<image>
Based on these two views showing the same scene: in which direction did I move from the first view to the second view? A. Diagonally forward and left B. Diagonally forward and right C. Directly right D. Directly left

Please reason step by step, and put your final answer (the option's letter) within \boxed{}.
\end{lstlisting}

\begin{lstlisting}[caption={Full-trace teacher prompt. The two leading image tokens are the original frames; the third is the agent's BEV ego-trajectory render.},label={lst:ft-teacher}]
<image>
<image>
Based on these two views showing the same scene: in which direction did I move from the first view to the second view? A. Diagonally forward and left B. Diagonally forward and right C. Directly right D. Directly left

Here is an execution trace from a vision-tool-augmented agent that analyzed the same input images with perception tools:
=== Agent Tool-Trace Begin ===
# Privileged Agent Reasoning Trace (SpatialClaw)
Below is the COMPLETE internal context of a tool-using spatial-reasoning agent (same base model) that solved THIS question correctly by writing & running Python in a Jupyter kernel with 3D-reconstruction / segmentation tools. Use it as privileged guidance.

## Execution Plan
### Task Analysis
This is a camera motion estimation problem requiring determination of ego-vehicle direction from two static frames. The implicit coordinate system is world-space (gravity-aligned), where +X=right, +Y=up, and camera forward=-Z direction. Target variable is movement direction vector between frames. Answer format is single selection from four directional options.

### Information Needs
1. 3D camera positions in world coordinates for both frames
2. Movement vector components (X for left/right, Z for forward/backward)
3. Visual confirmation of trajectory direction

### Computation Plan
1. **Reconstruct 3D scene** using both frames:
   ```
   recon = tools.Reconstruct.Reconstruct(InputImages[:2])
   ```
2. **Extract camera poses** (world coordinates):
   ```
   pos0 = recon.extrinsics[0][:3, 3]
   pos1 = recon.extrinsics[1][:3, 3]
   ```
3. **Compute movement vector**:
   ```
   mv = pos1 - pos0  # (dx, dy, dz)
   ```
4. **Determine direction** by analyzing components:
   - If dx > 0 and dz < 0 -> B (forward+right)
   - If dx < 0 and dz < 0 -> A (forward+left)
   - If dx > 0 and |dz| < 0.1 -> C (right)
   - If dx < 0 and |dz| < 0.1 -> D (left)
5. **Visual verification** via BEV rendering:
   ```
   recon.render_bev(ego_trajectory=True)
   ```
6. **Cross-validate** vector analysis with BEV trajectory direction

### Verification Checklist
```json
[
  {"priority": "HIGH", "description": "Verify reconstruction succeeded (extrinsics non-NaN, valid poses)"},
  {"priority": "HIGH", "description": "Confirm movement vector components have correct signs (dx for left/right, dz for forward)"},
  {"priority": "MEDIUM", "description": "Check BEV ego trajectory direction matches vector-based conclusion"},
  {"priority": "MEDIUM", "description": "Validate threshold for 'straight' vs 'diagonal' movement (|dz| < 0.1m)"},
  {"priority": "LOW", "description": "Sanity-check movement magnitude (pedestrian ~1-2m/s, car ~10-30m/s)"},
  {"priority": "LOW", "description": "Verify reconstruction quality via depth map visualization"}
]
```

### Fallbacks
- If reconstruction fails: Use SAM3 to segment stationary objects, compute pixel shifts, then convert to 3D using depth from Reconstruct
- If BEV trajectory unclear: Reconstruct with additional frames if available
- If vector ambiguity: Compare with VLM-based object motion analysis across frames

## Step-by-step Execution

### Step 0

**Purpose**: Reconstruct 3D scene geometry from the two input frames to obtain camera poses in world coordinates.

**Reasoning**: To determine the direction of movement, I need the 3D camera positions in world space. The Reconstruct tool will provide extrinsics (camera poses) for each frame, which I can use to compute the displacement vector between frame 0 and frame 1.

**Next Goal**: Extract camera positions from the reconstruction and compute the movement vector.

**Code**:
```python
recon = tools.Reconstruct.Reconstruct(InputImages[:2])

pos0 = recon.extrinsics[0][:3, 3]
pos1 = recon.extrinsics[1][:3, 3]

mv = pos1 - pos0
print(f"Frame 0 pos: {pos0}")
print(f"Frame 1 pos: {pos1}")
print(f"Movement vector (dx, dy, dz): {mv}")
```

**Execution Output**:
```
Frame 0 pos: [-6.67632621e-05  2.14540948e-01  1.16388564e-04]
Frame 1 pos: [-0.37664266  0.3066202  -0.42744154]
Movement vector (dx, dy, dz): [-0.3765759   0.09207925 -0.42755793]
```

### Step 1

**Purpose**: Analyze the movement vector components to determine the camera direction and visualize the trajectory for confirmation.

**Reasoning**: From the movement vector (dx=-0.38, dy=0.09, dz=-0.43): dx < 0 means moving LEFT, dz < 0 means moving FORWARD (since forward is -Z in world coordinates). This indicates diagonal movement forward and left. I'll verify this with the BEV ego trajectory visualization.

**Next Goal**: Render BEV with ego trajectory to visually confirm the movement direction, then submit the final answer.

**Code**:
```python
# Visual verification with ego trajectory
bev = recon.render_bev(ego_trajectory=True)
show(bev)

# Final direction determination
dx = mv[0]  # left/right
dz = mv[2]  # forward/backward (negative = forward)

print(f"dx (left/right): {dx:.3f}")
print(f"dz (forward): {dz:.3f}")

# Determine direction
if dx < 0 and dz < 0:
    answer = "A. Diagonally forward and left"
elif dx > 0 and dz < 0:
    answer = "B. Diagonally forward and right"
elif dx > 0 and abs(dz) < 0.1:
    answer = "C. Directly right"
elif dx < 0 and abs(dz) < 0.1:
    answer = "D. Directly left"

print(f"Answer: {answer}")
ReturnAnswer(answer)
```

**Execution Output**:
```
[show() visualization -> see image(s) below]
dx (left/right): -0.377
dz (forward): -0.428
Answer: A. Diagonally forward and left
[ReturnAnswer] Answer submitted: A. Diagonally forward and left
```

**show() visualization** (show()):

<image>

## Final Answer
Agent final answer: A. Diagonally forward and left
Ground-truth answer: A
=== Agent Tool-Trace End ===

After reading the agent's tool-trace above, make sure you truly understand what each measured fact looks like in the images --- do not copy or cite the trace itself. Now, verify the facts against the original images with your own eyes and, using your own words and independent reasoning, derive the answer to the question above. Think step by step, explore different interpretations, and don't be afraid to backtrack or re-examine the images if something doesn't work out.

Please reason step by step, and put your final answer (the option's letter) within \boxed{}.
\end{lstlisting}

\subsubsection{Version of Summary}

The Summary variant asks Qwen3.5-9B to distill the raw trace into a compact set of \emph{key reasoning points} plus \emph{decisive evidence}. Four rules define this protocol: keep only the final effective path, so failed or superseded tool calls and their images are dropped and no exploration is narrated; quote every tool-returned value verbatim, since the tool-free student can never recompute a metric readout; anchor every visual claim to the exact file path of the image that carries it, including agent-generated derived visualizations, which are the privileged artifacts the student cannot reproduce; and never state the answer, name a winning option, or pair a candidate with the words match/correct/answer. The teacher wrapper is the shared tool-trace wrapper of Listing~\ref{lst:trace-wrapper}, and the annotator user message is Listing~\ref{lst:sum-user}.

\begin{lstlisting}[caption={Qwen3.5-9B system prompt for the Summary variant (\texttt{style=full}). Reproduced verbatim.},label={lst:sum-sys}]
# Role
You are an expert annotator. Your job is to distill the trace of a tool-using spatial-reasoning agent into a COMPACT set of "privileged hints" for one sample.

# Background
- A visual-spatial question (often multiple-choice) was attempted by an agent that writes Python and calls vision tools (e.g. Reconstruct, SAM3, vlm.locate,  vlm.ask_with_thinking, show, Geometry) over several steps, observing each tool's output. In most cases the agent produced the CORRECT answer; the user message tells you explicitly when its final answer was WRONG.
- You are given (as TEXT): the question, the ground-truth answer, the list of the original input frames (already visible to the student), and the agent's privileged reasoning trace. Inside the trace, every agent-produced visualization image is written inline as a marker of the form `<<VIS IMAGE file=/abs/path.png>>` at the exact point where the agent saw it.
- Your output becomes PRIVILEGED CONTEXT used to train another model that will see ONLY the question + the original input frames. So keep exactly the information that would most help solve THIS specific question, and drop everything that is generic or not decisive.

# IGNORE (these are NOT key information)
- Any agent system prompt, tool documentation, API usage rules, coordinate-system tutorials, and generic "Execution Plan" boilerplate. They are identical across samples and carry no per-sample signal.
- Restatements of the question, response-format reminders, step/show() budget lines.
- Code mechanics (imports, variable names, print scaffolding) --- keep only the RESULTS the code produced.

# DISTILL THE TRACE TO ITS FINAL EFFECTIVE PATH (read carefully)
The trace is a raw exploration log and often contains detours. Before extracting anything, mentally reconstruct only the path that actually produced the correct answer, and DISCARD the rest. Specifically:
- If a tool call FAILED (error, empty mask, "Not visible", "Cannot determine",
  NaN, implausible value) and the agent retried or switched tools, KEEP ONLY the later successful attempt. Drop the failed attempt entirely.
- If the agent tried tool/approach A, found it unsatisfactory, and moved to tool/ approach B whose result it actually used, KEEP ONLY B. Do not mention A.
- If the agent repeated similar work to refine a result, keep ONLY the final, most reliable result --- not the intermediate tries.
- Do NOT narrate the exploration ("first tried X, it failed, then did Y"). Present only the clean, effective evidence as if the agent had gone straight to it.
- The ONLY exception: keep a one-line note about a discarded approach ONLY IF its failure is itself the decisive insight for the answer (e.g. "object X is not present in any frame" is the actual reason for the answer). Otherwise omit it.

# KEEP (the key information) --- as bullet points
1. Tool observations (final effective path only): the RESULTS that mattered --- e.g. the textual conclusion from vlm.ask_with_thinking, coordinates from vlm.locate, segmentation instance counts / mask areas, reconstruction-derived quantities (distances in meters, angles in degrees, left/right/front/behind dot-product outcomes, motion/trajectory findings). Preserve all numeric values and units VERBATIM. See "PRESERVE TOOL-RETURNED VALUES" below --- this is REQUIRED.
2. Visual evidence: the specific image(s) that carry the visual evidence the reasoning relies on --- both agent visualizations (BEV renders, mask overlays, depth maps, annotated frames) AND the original input frames the agent actually reasoned about. See "ANCHOR EVIDENCE TO IMAGES" below --- citing these is REQUIRED, not optional.
3. Reasoning chain: the minimal sequence of inferences and the verification method the agent used, presented as a clean straight line (no detours) --- but stopping SHORT of declaring the final choice (see the no-leak rule below).

# ANCHOR EVIDENCE TO IMAGES (required)
A core purpose of this hint is to DIRECT THE STUDENT'S ATTENTION to the right image. Therefore:
- Whenever a reasoning point depends on something seen in an image, you MUST anchor it to that specific image by citing its EXACT file path inline, then describing what to look at in it. Never describe visual evidence in the abstract ("the scene shows a kettle") without pointing to WHICH image shows it.
- Cite an agent visualization by copying its path verbatim from its `<<VIS IMAGE file=...>>` marker in the trace.
- Cite an original input frame by copying its exact path verbatim from the "Original input frames" list given in the user message (this re-shows that frame to focus attention on it).
- Prefer pointing to the actual image over only paraphrasing it. If several frames are compared, cite EACH frame you reference. Do NOT collapse the visual evidence into pure prose.
- Cite an image only when it genuinely carries evidence the reasoning uses; do not pad with images that play no role. You MAY refer back to an image you already cited (e.g. when first listing frames together, then describing each one) --- just reference it again by its exact path and it will reuse the same image handle.

# PRESERVE TOOL-RETURNED VALUES (required)
A key part of the privileged signal is the QUANTITATIVE / TEXTUAL data that tools computed and fed back to the VLM brain in their "Execution Output" --- the student, which has no tools, can never recompute these. Whenever a reasoning point relies on such a returned value, you MUST quote it VERBATIM (number + unit + what it refers to).
Examples of values to keep:
- camera / object positions and coordinates (e.g. "frames 0,1: X ~= -0.6 to -0.8 (camera moving); frames 2-5: X ~= 0 (stationary)"), trajectory / ordering results;
- distances in meters, heights, angles in degrees, depth values;
- left/right/front/behind or above/below outcomes and the dot-product / sign behind them; counts (instances, masks); areas; sizes; ratios;
- any explicit threshold, comparison, or numeric conclusion the agent acted on.
Do NOT paraphrase these into vague words ("the camera was roughly stable") when the trace gives concrete numbers --- keep the numbers. Attribute each value to the tool step that produced it only briefly (no code). Never invent values not in the trace.

# AGENT-GENERATED VISUALIZATIONS ARE HIGH-VALUE (always cite) 

Some images in the trace are agent-generated DERIVED visualizations --- BEV (bird's-eye-view) renders from render_bev, segmentation/mask overlays from seg.visualize, depth maps, annotated frames from draw_*, or plots from tools.Graph. These are privileged artifacts the student CANNOT reproduce, so they are especially valuable.
- You MUST cite EVERY derived visualization the agent generated and viewed (by its exact `<<VIS IMAGE file=...>>` path) and briefly say what it shows --- EVEN IF the agent's final textual conclusion leaned on the raw input frames instead. Do NOT drop a BEV / mask / depth / plot just because the decisive text reasoning used the original frames.
- Recognize a derived visualization from the code that produced it (render_bev, .visualize(, draw_bbox/point/line, depth, plot, tools.Graph). A plain show(InputImages) that only re-displays the original frames is NOT derived --- those are echoes of the input frames, cite them only when an input frame is itself the decisive evidence.

# DROP
- All failed, abandoned, or superseded tool calls and the rework they caused, INCLUDING any image produced by such a discarded step.
- Duplicated evidence --- keep only the strongest / final instance. (Referring back to the same image more than once is fine and reuses its handle; this is about not keeping redundant SEPARATE observations.)

# IF THE AGENT'S FINAL ANSWER WAS WRONG (the user message will say so explicitly) 

Some traces end in a WRONG final answer. For those samples:
- The ground-truth answer given in the user message is the ONLY reliable target; the agent's final choice and the reasoning specific to that choice are noise.
- Reconstruct the subset of tool observations that are CONSISTENT with the ground-truth answer (tool measurements are often valid even when the agent's final reading of them is not) and present those as the key reasoning points, as if the agent had followed the correct path.
- DROP the agent's erroneous conclusions, its final-answer declaration, and any inference step that only makes sense for the wrong answer.
- If a tool result itself contradicts the ground truth (e.g. a mismeasured direction), DROP that result too --- never include evidence that would lead the student toward the wrong answer.
- All other rules (no-leak, image anchoring, verbatim values) apply unchanged.

# DO NOT LEAK THE ANSWER (hard rule)
The ground-truth answer is given to you ONLY so you can identify the effective reasoning path. Your output is a HINT, not the solution --- the student must still derive the answer itself. Therefore:
- NEVER state the final answer in any form: no option letter (A/B/C/D), no number, no free-form answer value, no "Final Answer", no "the answer is ...".
- NEVER declare which candidate/option/image is the correct one (e.g. do NOT write "Option D matches", "Image 5 is the answer", "B is the correct view"). Do not pair a candidate with the word match/correct/answer.
- Instead, describe the DISTINGUISHING EVIDENCE and the features to compare (e.g. "the egocentric frame that shows a black tool cart with red rags, a silver tripod on the background left, and the bike wheel on the right is the one consistent with the exocentric layout"), phrased so the reader must still perform the final matching/decision themselves.
- Refer to images by neutral descriptors or their file path, NOT by their answer label. It is fine to describe what each candidate shows; it is NOT fine to say which candidate wins.

# IMAGE / VIDEO PATH FIDELITY (hard rule)
- Whenever your output references a visualization image, you MUST copy its file path EXACTLY as it appears inside the `<<VIS IMAGE file=...>>` marker --- character for character (full path, including directories and extension).
- NEVER rename, shorten, paraphrase, reformat, change extension, renumber, or otherwise alter any path. Do not invent paths. Only cite paths that actually appear in the trace.

# Output format (output ONLY this, no preamble, no markdown headers)
Key reasoning points:
- <one self-contained fact/observation per bullet, in the order it becomes relevant; anchor every visual claim by citing the exact file path of the image that shows it (visualization or original input frame)>
- ...
Decisive evidence:
- <the 1-2 cues that distinguish the correct interpretation, each anchored to the exact image path that shows it, WITHOUT stating the final answer or naming the winning option/candidate>

# Constraints
- Be faithful: do NOT invent evidence that is not in the trace.
- Do NOT include any generic tool/usage knowledge.
- Do NOT leak the answer (see the no-leak rule above): no answer value, no winning option letter, no "this candidate is the match".
- ALWAYS anchor visual evidence to a specific image path (see "ANCHOR EVIDENCE TO IMAGES"); do not reduce the hint to pure prose.
- Be concise: the smallest set of bullets that fully captures the evidence and the method that make the answer derivable. Preserve numbers/units AND file paths exactly.
\end{lstlisting}

\paragraph{OPSD example.} Same instance as Listing~\ref{lst:ft-student}. The annotator cited three images, in this order: original frame~1, original frame~2, and the BEV render. Citation tokenization therefore turns the two leading citations into the handles \texttt{(image 1)} and \texttt{(image 2)} and the BEV into \texttt{(image 3)}. The student receives $2$ images; the teacher receives $5$, namely the two original frames at training resolution followed by the three cited images, the first two of which are $400$\,px thumbnails of the same two frames.

\begin{lstlisting}[caption={Summary student prompt.},label={lst:sum-student}]
<image>
<image>
Based on these two views showing the same scene: in which direction did I move from the first view to the second view? A. Diagonally forward and left B. Diagonally forward and right C. Directly right D. Directly left

Please reason step by step, and put your final answer (the option's letter) within \boxed{}.
\end{lstlisting}

\begin{lstlisting}[caption={Summary teacher prompt. The privileged body preserves the metric readouts verbatim and names no option letter.},label={lst:sum-teacher}]
<image>
<image>
Based on these two views showing the same scene: in which direction did I move from the first view to the second view? A. Diagonally forward and left B. Diagonally forward and right C. Directly right D. Directly left

Here is an execution trace from a vision-tool-augmented agent that analyzed the same input images with perception tools:
=== Agent Tool-Trace Begin ===
Key reasoning points:
- 3D reconstruction of the two input frames (`<image> (image 1)` and `<image> (image 2)`) yields camera positions: Frame 0 at approximately (0, 0.21, 0) and Frame 1 at approximately (-0.38, 0.31, -0.43).
- The computed movement vector between the frames is (dx, dy, dz) = (-0.377, 0.092, -0.428).
- In the world coordinate system where +X is right and -Z is forward, the negative dx (-0.377) indicates movement to the left, and the negative dz (-0.428) indicates movement forward.
- The bird's-eye-view (BEV) render with the ego trajectory visually confirms the trajectory direction corresponding to these negative dx and dz components.

Decisive evidence:
- The movement vector components dx = -0.377 (left) and dz = -0.428 (forward) derived from the 3D reconstruction of the two frames.
- The BEV visualization (`<image> (image 3)`) showing the ego trajectory moving in the direction indicated by the negative dx and dz components.
=== Agent Tool-Trace End ===

After reading the agent's tool-trace above, make sure you truly understand what each measured fact looks like in the images --- do not copy or cite the trace itself. Now, verify the facts against the original images with your own eyes and, using your own words and independent reasoning, derive the answer to the question above. Think step by step, explore different interpretations, and don't be afraid to backtrack or re-examine the images if something doesn't work out.

Please reason step by step, and put your final answer (the option's letter) within \boxed{}.
\end{lstlisting}

\subsubsection{Version of Fact}

The Fact variant isolates the \emph{perceptual} content of the trace from its \emph{inferential} content. Its governing rule is a recoverable-privilege criterion: effective privilege should be a visual cue whose source is fully contained in the input frames, i.e.\ something a perfect observer could re-derive by inspecting the images alone, so that the student can learn to see it rather than to memorize it. Answer-side content --- the final choice, the spatial relation the question asks about, a matching or ordering verdict, or the solving idea --- is not recoverable from the images by definition of the task, and supervising on it trains shortcut imitation. Concretely, raw pose numbers and axis conventions are kept verbatim while their verbalization into direction words is dropped; per-frame content observations are kept while cross-frame verdicts and temporal orderings are dropped; and every bullet is stated declaratively in perception order, with no inference connectives and no ranking of facts as decisive. The teacher wrapper is again Listing~\ref{lst:trace-wrapper} and the annotator user message is Listing~\ref{lst:sum-user}.

\begin{lstlisting}[caption={Qwen3.5-9B system prompt for the Fact variant (\texttt{style=facts}). Reproduced verbatim.},label={lst:fact-sys}]
# Role
You are an expert annotator. Your job is to extract from the trace of a tool-using spatial-reasoning agent ONLY the tool-measured PERCEPTUAL FACTS for one sample --- NOT the agent's reasoning or conclusions.

# Background
- A visual-spatial question (multiple-choice) was attempted by an agent that writes Python and calls vision tools (e.g. Reconstruct, SAM3, vlm.locate, vlm.ask_with_thinking, show, Geometry) over several steps. In most cases it produced the CORRECT answer; the user message tells you explicitly when its final answer was WRONG.
- You are given (as TEXT): the question, the ground-truth answer, the list of the original input frames (already visible to the student), and the agent's privileged reasoning trace. Inside the trace, every agent-produced visualization image is written inline as a marker of the form `<<VIS IMAGE file=/abs/path.png>>` at the exact point where the agent saw it.
- Your output becomes PRIVILEGED CONTEXT for training another model that sees ONLY the question + the original input frames.

# RECOVERABLE-PRIVILEGE CRITERION (the one rule everything follows from) Effective privilege must be a VISUAL CUE: a description of question-relevant evidence whose SOURCE is fully contained in the input frames --- i.e. something a perfect observer could re-derive by inspecting the images alone (S = f(images, question)). Such content adds no information beyond the student's own input, so the student can learn to "see" it instead of memorizing it. Answer-side content (the final choice, the spatial relation the question asks about, a matching or ordering verdict, the solving idea) is NOT recoverable from the images by definition of the task --- supervising on it trains shortcut imitation. Test every bullet you write against this criterion:
- "cam1 at [0.82, 0.03, -0.41] in cam0 frame (x=right, y=up, z=forward)" --- a measurement OF the images. KEEP.
- "therefore camera 1 is to the left-rear" --- the question's answer restated in words. DROP (keep the numbers it came from).
- "the diagram shows two intersecting chords with segments x, 8, 4, 6" --- visual structure. KEEP. / "apply the intersecting-chords theorem" --- solving idea. DROP.

# KEEP --- tool-measured perceptual facts ONLY (quote values VERBATIM)
1. Reconstruction outputs: camera positions / orientations / pose matrices and coordinates (e.g. "cam0 at origin, cam1 at [0.82, 0.03, -0.41] in cam0 frame, axes: x=right, y=up, z=forward"), translation/displacement vectors, per-frame trajectory coordinates, distances in meters, angles in degrees, depth values. KEEP THE RAW NUMBERS AND AXIS CONVENTIONS --- they are measurements, not conclusions. Do NOT verbalize them into direction words.
2. Correspondence / matching measurements: which object/feature/landmark is visible in which frame and where (pixel boxes/points from vlm.locate, SAM3 instance counts, mask areas), relative depth or apparent-size readings of the SAME object across frames.
3. Per-frame content observations from tools (vlm.ask_with_thinking / vlm.locate textual outputs describing what IS visible in a single frame: objects, states, attributes, layout WITHIN that frame).
4. Agent-generated derived visualizations (BEV renders, mask overlays, depth maps, annotated frames, plots): cite EVERY one by its exact `<<VIS IMAGE file=...>>` path and describe ONLY what it depicts (e.g. "top-down BEV of the two camera frusta and the sofa point cloud") --- without reading a direction/answer off it.
5. Original input frames that carry decisive visual evidence: cite by their exact path from the "Original input frames" list.

# DROP --- everything conclusion-shaped (this is the core rule)
- ALL directional conclusions about cameras/viewpoints: any sentence like "therefore/so/this means/indicating the camera (1/2) is to the LEFT-REAR / moved RIGHT-FORWARD / is behind ...". The raw pose/displacement NUMBERS are kept (see KEEP-1); their verbalization into left/right/front/back words is NOT.
- Any statement pairing a candidate frame/option with match/correctness/answer ("frame B shows the same cart", "the exo view corresponds to candidate 2"). Per-frame content facts are fine; the cross-frame VERDICT is not.
- Any temporal-order conclusion ("the order is D-A-C-B", "frame 2 comes first"). Per-frame state measurements (door half-open, object moved to pixel X) are fine.
- The agent's reasoning chain, justifications, comparisons that resolve the question, verification narration, and "decisive evidence" style hints.
- All failed/abandoned/superseded tool calls and their outputs or images; keep only the final effective measurement when the agent retried or refined.
- Agent system prompt, tool docs, plans, restatements of the question, code mechanics. Only the RESULTS tools returned matter.

# IF THE AGENT'S FINAL ANSWER WAS WRONG (the user message will say so explicitly) Some traces end in a WRONG final answer. For those samples:
- The ground-truth answer given in the user message is the ONLY reliable target; the agent's final choice is noise.
- Keep only the tool measurements that are CONSISTENT with the ground-truth answer. A measurement that itself contradicts the ground truth (e.g. a mismeasured pose or a wrong per-frame observation that drove the agent's error) must be DROPPED --- never include facts that would lead a student reasoning correctly over them toward the wrong answer.
- All other rules (recoverable-privilege criterion, no-leak, neutrality, verbatim values, path fidelity) apply unchanged.

# DO NOT LEAK THE ANSWER (hard rule, strengthened)
The ground-truth answer is given ONLY so you can identify which measurements were actually load-bearing. Your output must NOT contain:
- the final answer in any form (no option letter A/B/C/D, no answer phrase);
- any direction word (left/right/forward/backward/front/back/behind, or their  combinations) used as a CONCLUSION about camera/viewpoint relations or motion;
- any "X matches / corresponds to / is the correct ..." verdict about a candidate;
- any stated ordering of frames for ordering questions. If a tool's textual output itself contains such a conclusion sentence (e.g. a vlm.ask answer saying "the camera moved left"), DROP that sentence; keep only the measurement parts around it. Direction words are allowed ONLY inside axis conventions ("x=right") or within-frame object layout ("the sink is left of the stove in frame 1"), NEVER for the camera-to-camera / frame-to-frame relation the question asks about.

# IMAGE / VIDEO PATH FIDELITY (hard rule)
- Copy every cited file path EXACTLY as it appears (inside `<<VIS IMAGE file=...>>` or the original-frames list) --- character for character.
- NEVER rename, shorten, reformat, renumber, or invent paths.

# PRESENT FACTS NEUTRALLY (no argument structure)
- State every bullet DECLARATIVELY, in perception order (frame by frame / tool by tool), NOT arranged as an argument building to a conclusion.
- No inference connectives: never "therefore / thus / so / hence / this means / which indicates / confirming that".
- No instructions or strategy hints: never "compare X with Y", "check whether", "to determine the answer, ...", "note that this distinguishes ...".
- Do not rank or highlight facts as "decisive/key/critical" --- list them flat.

# QUALITY CHECKLIST (all three must hold)
1. COMPLETE: every measurement the agent's successful path actually relied on is present (missing question-relevant evidence makes the privilege useless).
2. VISUAL-ONLY: nothing answer-side slipped in (no verdict, no direction conclusion, no ordering, no strategy) and nothing non-visual was added.
3. UNAMBIGUOUS: every fact is anchored to its exact image path and quotes exact numbers/units --- no vague paraphrases ("roughly stable", "quite close").

# Output format (output ONLY this, no preamble, no markdown headers)
Tool-measured perceptual facts:
- <one measurement/observation per bullet, numbers+units+axis conventions verbatim, each visual claim anchored to the exact file path of the image that shows it>
- ...
Derived visualizations:
- <exact path> --- <what this visualization depicts, perceptual description only>

# Constraints
- Be faithful: do NOT invent measurements not present in the trace.
- Every bullet must be a FACT (measurement/observation), never an inference that resolves the question. When in doubt, keep the numbers, drop the words.
- Be concise: the smallest set of facts that makes the answer derivable by a student who reasons over them; preserve numbers/units AND file paths exactly.
\end{lstlisting}

\paragraph{OPSD example.} Same instance. The annotator cited a single image, the BEV render, so the teacher receives $3$ images. The \texttt{\#\# Final Answer} section is the deterministic post-hoc append described in the overview, not annotator output; the fact list itself contains no directional conclusion.

\begin{lstlisting}[caption={Fact student prompt.},label={lst:fact-student}]
<image>
<image>
Based on these two views showing the same scene: in which direction did I move from the first view to the second view? A. Diagonally forward and left B. Diagonally forward and right C. Directly right D. Directly left

Please reason step by step, and put your final answer (the option's letter) within \boxed{}.
\end{lstlisting}

\begin{lstlisting}[caption={Fact teacher prompt. Pose and displacement numbers are quoted verbatim; the axis convention is retained, while ``left'' and ``forward'' as camera-motion conclusions are absent.},label={lst:fact-teacher}]
<image>
<image>
Based on these two views showing the same scene: in which direction did I move from the first view to the second view? A. Diagonally forward and left B. Diagonally forward and right C. Directly right D. Directly left

Here is an execution trace from a vision-tool-augmented agent that analyzed the same input images with perception tools:
=== Agent Tool-Trace Begin ===
Tool-measured perceptual facts:
- World-space coordinate system convention: +X=right, +Y=up, camera forward=-Z direction
- Frame 0 camera position in world coordinates: [-6.67632621e-05, 2.14540948e-01, 1.16388564e-04]
- Frame 1 camera position in world coordinates: [-0.37664266, 0.3066202, -0.42744154]
- Movement vector from Frame 0 to Frame 1 (dx, dy, dz): [-0.3765759, 0.09207925, -0.42755793]

Derived visualizations:
- <image> (image 1) --- BEV rendering with ego trajectory

## Final Answer
Agent final answer: A. Diagonally forward and left
Ground-truth answer: A
=== Agent Tool-Trace End ===

After reading the agent's tool-trace above, make sure you truly understand what each measured fact looks like in the images --- do not copy or cite the trace itself. Now, verify the facts against the original images with your own eyes and, using your own words and independent reasoning, derive the answer to the question above. Think step by step, explore different interpretations, and don't be afraid to backtrack or re-examine the images if something doesn't work out.

Please reason step by step, and put your final answer (the option's letter) within \boxed{}.
\end{lstlisting}

\subsubsection{Version of Intent}

The Intent variant recovers, for each effective tool call, the \emph{reason} the agent made that call --- the one piece of visual information it needed to advance --- and rewrites it as a tool-free instruction a reader can execute by looking at the images. A call such as \texttt{Reconstruct() -> cam1 at [0.82, 0.03, -0.41] relative to cam0} becomes ``judge how the viewpoint shifted between image~1 and image~2 by tracking how fixed landmarks move across the two frames''. The privileged body is thus a transferable skill rather than a worked solution: it states what to do, never the result of doing it, and it may state generic visual-reasoning principles (for instance, that static objects appear to shift left in the frame when the camera translates right), which are exactly the transferable part. Tool names, tool-returned values, file paths, and any statement of the answer are banned. Frames are referred to only as ``image~1''\,\ldots\,``image~$N$''. Because the sanitizer removes any surviving path or visualization reference, the teacher sees exactly the student's frames and no extra image, which makes Intent the variant with the narrowest privilege channel: the teacher's advantage is purely procedural.

\begin{lstlisting}[caption={Qwen3.5-9B system prompt for the Intent variant (\texttt{style=intent}). Reproduced verbatim.},label={lst:int-sys}]
# Role
You are an expert annotator. Your job is to distill the trace of a tool-using spatial-reasoning agent into a compact, TOOL-FREE REASONING STRATEGY --- a transferable "skill" for one sample: the way of looking at and thinking about the input images that leads to the answer of THIS question, executable with eyes alone.

# Background
- A visual-spatial question (usually multiple-choice) was attempted by an agent that writes Python and calls vision tools (3D reconstruction, segmentation, VLM queries, visualization) over several steps, observing each tool's output. The user message tells you explicitly whether its final answer was CORRECT or WRONG.
- You are given (as TEXT): the question, the ground-truth answer, the ordered list of the original input frames (the only images the student will see), and the agent's full step-by-step trace. Inside the trace, agent-produced visualization images appear inline as `<<VIS IMAGE file=...>>` markers --- they are the agent's private tool artifacts and must never surface in your output.
- Your output becomes PRIVILEGED CONTEXT used to train another model that sees ONLY the question + the original input frames and has NO tools. It will be introduced to that model as "a reasoning strategy distilled from the trace of a tool-using agent", so write it as direct, fluent guidance to the reader ("First, ...", "Then, compare ...") --- never as a report about what some agent did.

# CORE TRANSFORMATION: from tool calls to intents
The trace is a sequence of tool calls. Each effective tool call was made for a REASON: the agent needed one specific piece of visual information to advance its reasoning. Recover that chain of intents and rewrite each one as a visual-reasoning instruction that a tool-less reader can carry out by looking at the images:
- trace: "vlm.locate('kettle', frame 2) -> bbox [412, 88, ...]" strategy: "Locate the kettle in image 2 and note which side of the counter it sits on."
- trace: "Reconstruct() -> cam1 at [0.82, 0.03, -0.41] relative to cam0" strategy: "Judge how the viewpoint shifted between image 1 and image 2 by tracking how fixed landmarks (the counter edge, the window) move across the two frames."
- trace: "SAM3('chair') -> 6 instances" strategy: "Count the chairs around the table, including partially occluded ones."
Keep the INSTRUCTION (what to look at, what to compare, what to infer, and why this step helps); drop the tool, the mechanics, and the returned value.

# HARD BAN: no tools, no tool results
The strategy must read as if written by an expert who solved the question by careful looking and thinking alone. Therefore NEVER mention:
- tools, code, Python, function or tool names, "the agent", "the trace", reconstruction, segmentation, masks, bounding boxes, pixel coordinates, depth maps, BEV / bird's-eye-view renders, plots, or any other tool artifact;
- any tool-returned VALUE: no 3D coordinates, no distances in meters, no angles in degrees, no pixel positions, no instance counts, no areas, no numeric thresholds read off a tool output;
- any `<<VIS IMAGE ...>>` marker or file path. State WHAT TO DO ("estimate the relative distance of the sofa in both views by comparing its apparent size"), never the RESULT of doing it ("the sofa is closer in the second view"). Observations and conclusions belong to the reader. Generic visual-reasoning principles (e.g. "when the camera translates right, static objects appear to shift left in the frame") are allowed and encouraged where they underpin a step --- they are exactly the transferable part of the skill.

# DISTILL THE FINAL EFFECTIVE PATH ONLY
The trace is a raw exploration log and often contains detours. Reconstruct only the path that actually produced (or, for wrong-answer traces, would have produced) the
correct answer:
- Drop failed, abandoned, or superseded attempts entirely; do not narrate the exploration ("first X was tried, then ...").
- Merge repeated refinements of the same measurement into the single step they amount to.
- Exception: if a failure IS the decisive insight (e.g. the object the question asks about is absent from every frame), turn it into a positive checking step ("scan each frame for X and note whether it appears at all").

# IF THE AGENT'S FINAL ANSWER WAS WRONG (the user message says so explicitly)
- The ground-truth answer (given to you) defines which path counts as "effective". Keep the intents of steps that gather evidence consistent with the ground truth; DROP steps whose only purpose was to support the wrong conclusion, and steps the agent executed in a misleading way.
- If the agent went wrong because it overlooked or misread something, ADD a caution step at the right place: what to examine extra carefully and which mistake to avoid --- phrased generically, without revealing the correct conclusion.

# DO NOT LEAK THE ANSWER (hard rule)
The ground-truth answer is given to you ONLY so you can identify the effective path.
The strategy must make the answer DERIVABLE, never SPELLED OUT:
- NEVER state the final answer in any form: no option letter (A/B/C/D), no answer phrase, no "the correct option/order/direction is ...".
- NEVER state as a fact the very relation/quantity/ordering the question asks about (for "which direction are you moving?" do NOT write "the camera moves forward-left" --- write HOW to determine the direction from the images).
- NEVER pair a candidate/option/frame with words like "correct", "matches", "the answer". Describing what distinguishes the candidates is fine; naming the winner is not.

# FRAME REFERENCES
Refer to the input frames ONLY as "image 1" ... "image N" (or "the first image", "the second image"), following the numbering of the "Original input frames" list in the user message. Never use file paths. Naming concrete objects and landmarks visible in the images ("the red sofa", "the kitchen island") is encouraged --- it directs attention without leaking results.

# Output format (output ONLY this, no preamble, no markdown headers)
Reasoning strategy:
1. <one step per intent, in execution order: what to look at / compare / infer, and what this step establishes for the reasoning>
2. ...Decision criterion:
- <how the accumulated observations discriminate between the candidate answers --- the distinguishing evidence to weigh, WITHOUT stating which candidate wins>

# Constraints
- Faithful: the strategy must follow the approach actually present in the trace (with ground-truth-guided corrections for wrong-answer traces); do not invent a different method.
- Self-contained and fluent: direct imperative guidance; no meta-talk about agents, tools, traces, or annotation.
- Concise: the smallest number of steps that fully transfers the approach --- typically 3-8 steps.
\end{lstlisting}

\begin{lstlisting}[caption={Qwen3.5-9B user message for the Intent variant. The frame list is numbered so that the annotator can only refer to frames as ``image~$k$'', never by path.},label={lst:int-user}]
## Question
{question}

## Ground-truth answer
{ground_truth_answer}

## Agent outcome
The agent's final answer was CORRECT.

## Original input frames (refer to them ONLY as "image 1" ... "image N" in this order; NEVER by file path)
  image 1 = {absolute_path_of_original_frame_1}
  image 2 = {absolute_path_of_original_frame_2}
  ...

## Privileged agent reasoning trace
(Visualization images appear inline as <<VIS IMAGE file=...>> --- they are the agent's private tool artifacts; NEVER cite or mention them in your output.)

{raw privileged context, with the k-th <image> token replaced by <<VIS IMAGE file={show_images[k]}>>}
\end{lstlisting}

\begin{lstlisting}[caption={Teacher-side wrapper for the Intent variant. Reproduced verbatim.},label={lst:intent-wrapper}] {question core, i.e. <image> tokens + question + options}

Below is a reasoning strategy for this question. It is experience distilled from the execution trace of a vision-tool-augmented agent that analyzed the same input images with perception tools: each step condenses the intent behind one of the agent's tool calls into a step of pure visual reasoning, so the whole strategy can be carried out just by looking at the images --- no tools required:
=== Distilled Reasoning Strategy Begin ===
{privileged body}
=== Distilled Reasoning Strategy End ===

The strategy above tells you what to look for and how to reason --- it is not a worked-out solution, and it deliberately contains no measurements or conclusions. Follow its steps with your own eyes on the original images above: make every observation yourself, and combine what you actually see into your own chain of reasoning to derive the answer to the question. Think step by step, explore different interpretations, and don't be afraid to backtrack or re-examine the images if something doesn't work out.

Please reason step by step, and put your final answer (the option's letter) within \boxed{}.
\end{lstlisting}

\paragraph{OPSD example.} Same instance. Student and teacher each receive the same $2$ images. Note how the metric displacement of the original trace has been replaced by the parallax and apparent-size cues that produce the same judgement from the pixels alone, and how the decision criterion states the discriminating cues without naming the winning option.

\begin{lstlisting}[caption={Intent student prompt.},label={lst:int-student}]
<image>
<image>
Based on these two views showing the same scene: in which direction did I move from the first view to the second view? A. Diagonally forward and left B. Diagonally forward and right C. Directly right D. Directly left

Please reason step by step, and put your final answer (the option's letter) within \boxed{}.
\end{lstlisting}

\begin{lstlisting}[caption={Intent teacher prompt. No privileged image, no metric readout, and no stated answer.},label={lst:int-teacher}]
<image>
<image>
Based on these two views showing the same scene: in which direction did I move from the first view to the second view? A. Diagonally forward and left B. Diagonally forward and right C. Directly right D. Directly left

Below is a reasoning strategy for this question. It is experience distilled from the execution trace of a vision-tool-augmented agent that analyzed the same input images with perception tools: each step condenses the intent behind one of the agent's tool calls into a step of pure visual reasoning, so the whole strategy can be carried out just by looking at the images --- no tools required:
=== Distilled Reasoning Strategy Begin ===
Reasoning strategy:
1. Identify several distinct, stationary landmarks or objects at varying depths (e.g., foreground and background) in image 1.
2. Determine the forward/backward component by comparing the apparent size and depth of these landmarks in image 2. If the objects appear larger and the viewpoint has advanced deeper into the scene, the movement has a forward component.
3. Determine the left/right component by observing the lateral parallax and the visible faces of the objects. If the scene appears to shift rightward relative to the frame, and you can see more of the right sides of objects while their left sides become more occluded, the movement has a leftward component.
4. Synthesize the depth and lateral observations to determine whether the movement is purely along one axis or diagonal, and in which specific direction.

Decision criterion:
- A forward component is established by objects appearing larger and closer in the second image. A leftward component is established by objects shifting rightward in the frame and revealing more of their right sides. The combination of these two visual cues identifies the specific diagonal direction, distinguishing it from purely lateral or other diagonal movements.
=== Distilled Reasoning Strategy End ===

The strategy above tells you what to look for and how to reason --- it is not a worked-out solution, and it deliberately contains no measurements or conclusions. Follow its steps with your own eyes on the original images above: make every observation yourself, and combine what you actually see into your own chain of reasoning to derive the answer to the question. Think step by step, explore different interpretations, and don't be afraid to backtrack or re-examine the images if something doesn't work out.

Please reason step by step, and put your final answer (the option's letter) within \boxed{}.
\end{lstlisting}

\subsection{Training data construction and baseline setup}
\label{app:data}

We run SpatialClaw with a Qwen3.5-9B backbone on the full MindCube training split of 10,000 multi-view spatial questions. The agent answers 5,008 questions correctly (50.08\%), and these ground-truth-verified instances form the shared training set for SpatialOPSD and GRPO. SpatialOPSD's privileged-information records are constructed from the corresponding verified agent traces, and all privileged-information variants use the same 5,008 instances. The SFT dataset is constructed separately: Qwen3.7-plus generates one tool-free rationale for each of the 10,000 training instances, and we retain the 4,459 instances whose final answers match the ground-truth labels. Thus, the SFT training set is not matched sample-for-sample with the SpatialOPSD and GRPO training set. Appendix A.2 details the construction of the privileged-information records.

Shared input format. For every instance, all methods receive byte-identical student inputs: the original frames resized to a maximum long edge of 2,048 px, the question and options with all dataset-specific direct-answer instructions stripped, and the chain-of-thought suffix "Please reason step by step, and put your final answer (the option's letter) within \verb|\boxed{}|" (full construction in Appendix A.2, Stage 4). All methods are trained to emit the final answer in the same \verb|\boxed{}| format, and all methods are evaluated under identical decoding settings.

% \paragraph{SFT baseline.} Each SFT record pairs the shared student input above with a tool-free chain-of-thought rationale written by Qwen3.7-plus. The rationale generator is conditioned on exactly the student-visible input — no tool outputs, no privileged information, and no frames other than the original input frames — and every rationale ends with the answer in \verb|\boxed{}|. For each instance we sample k = 8 rationales at temperature 0.7 and keep those whose extracted answer matches the ground-truth label; one rationale per instance is retained (the shortest correct one). Qwen3.5-9B produces a correct rationale for 5,008 instances; the 0 instances it misses are removed from the training sets of all methods, so SFT and SpatialOPSD remain matched sample-for-sample (final n = 5,008). The loss is computed on assistant tokens only. 
\paragraph{SFT baseline.}

We fine-tune Qwen3.5-9B for one epoch on the 4{,}459 retained MindCube instances using supervised next-token cross-entropy. The global batch size is 16 questions, with one target response per question. We use AdamW with a peak learning rate of $1\times10^{-5}$, a cosine schedule decaying to zero, and 9 warmup steps. Training uses eight GPUs with FSDP2, bf16 mixed precision, gradient checkpointing, and gradient accumulation over two steps. The loss is averaged over supervised assistant tokens.

% GRPO baseline. GRPO is trained on the same 5,000 student prompts, with the same image preprocessing and the same $\text{boxed{}}$ answer format. Its reward is an exact match between the extracted \boxed{} answer and the ground-truth label. 

\paragraph{GRPO baseline.}
Starting from Qwen3.5-9B, we run GRPO for one epoch over 5{,}008
spatial-reasoning questions, which gives 156 update steps. Each step samples a
group of $G=8$ responses per question for a batch size of $32$ questions
($32\times 8=256$ sequences per step); the whole batch forms a single mini-batch,
so exactly one gradient update is taken per rollout batch and training stays
strictly on-policy. Responses are capped at a maximum length of $4{,}096$ tokens
and prompts at $14{,}336$ tokens, with at most 12 frames per sample (long edge
$\leq 2048$\,px), sampled at temperature $1.0$ and top-$p=0.9$. We optimise with
AdamW ($\beta_1{=}0.9$, $\beta_2{=}0.999$, weight decay $0.01$, gradient clipping
$1.0$) at a learning rate of $1\times 10^{-6}$, warmed up over 10 steps and then
annealed by a cosine schedule to $10\%$ of its peak value. The reward is the
binary correctness of the extracted multiple-choice answer, advantages are the
group-relative rewards normalised by the group standard deviation, and the
policy-gradient ratio is clipped at $\epsilon=0.2$; we use no KL penalty to a
reference policy and no entropy bonus. Training runs on a single node of 8 GPUs
with FSDP (parameter and optimizer CPU offloading), Ulysses sequence parallelism
of degree 4, gradient checkpointing, bf16 mixed precision, and vLLM for
generation, and the rollout--training probability mismatch is corrected with
token-level truncated importance sampling (clip $2.0$).

\paragraph{SpatialOPSD.}
OPSD is trained from the same Qwen3.5-9B initialisation, on the same 5{,}008
questions for one epoch (156 update steps), and on the same 8-GPU FSDP/vLLM
stack with the same importance-sampling correction, so the only differences are
in the objective and the quantities it requires. It needs no reward, advantage,
or value model: each step draws a single response per question ($G=1$) at
temperature $1.0$ and top-$p=0.9$, giving $32$ sequences for a batch size of $32$
questions, again in one mini-batch and hence one on-policy update per step.
Responses are capped at a maximum length of $2{,}048$ tokens and the student
prompt at $14{,}336$ tokens, while the teacher additionally reads the privileged
context within an $18{,}432$-token budget. The loss is a token-level reverse KL
($\alpha=1$) to that frozen teacher, truncated to the student's top-$100$
support plus one tail bucket, together with a repetition control term
($4$-grams, minimum run $32$, unlikelihood coefficient $0.1$) that suppresses
verbatim transcription of the privileged block. The learning rate is a constant
$1\times 10^{-6}$ after 5 warmup steps, with the same AdamW settings, no KL
penalty to a reference policy, and no entropy bonus.

\subsection{Privileged Information Leakage Measurement}
\label{app:privileged_information_leakage}

We quantify privileged information leakage using a deliberately loose,
keyword-based criterion. The criterion detects whether a generated
response contains at least one standalone occurrence of any of three
predefined leakage keywords: \texttt{agent}, \texttt{trace}, or
\texttt{tool}. This metric is intended to measure surface-form disclosure
under a reproducible lexical rule; it does not attempt to infer the
semantic intent of a matched occurrence.

\subsubsection{Keyword Matching Rule}

For each keyword
\(w \in \{\texttt{agent}, \texttt{trace}, \texttt{tool}\}\),
matching is performed with the following case-insensitive regular
expression:

\begin{verbatim}
\b{w}s?(?:['\u2019]s)?\b
\end{verbatim}

Equivalently, the combined expression used to determine whether any
leakage keyword is present is:

\begin{verbatim}
\b(?:agent|trace|tool)s?(?:['\u2019]s)?\b
\end{verbatim}

Here, \verb|\b| requires a transition between a word character and a
non-word character on both sides of the matched form; \verb|s?| permits
an optional plural suffix; and \verb|(?:['\u2019]s)?| permits an optional
apostrophe-plus-\texttt{s} possessive suffix written with either a
straight apostrophe (\verb|'|) or a typographic apostrophe
(\textquoteright). Matching is case-insensitive.

Consequently, the detector recognizes singular, plural, and possessive
forms such as \texttt{agent}, \texttt{agents}, \verb|agent's|,
\texttt{trace}, \texttt{traces}, \texttt{tool}, \texttt{tools}, and
\verb|tool's|, regardless of capitalization. A plural possessive such as
\verb|agents'| is also classified as a hit because the substring
\texttt{agents} forms a complete match before the trailing apostrophe.
A keyword embedded within a longer alphanumeric word is not matched.
Thus, terms such as \texttt{toolkit}, \texttt{toolbar}, \texttt{toolbox},
\texttt{traceable}, \texttt{traced}, \texttt{tracing}, \texttt{agency},
\texttt{agenda}, and \texttt{reagent} are excluded. By contrast, a hyphen
or similar non-word character adjacent to a target form creates the
required boundary, so constructions such as \texttt{tool-trace} and
\texttt{tool-augmented} produce matches.

The criterion is lexical rather than semantic. A standalone keyword is
counted even when used in an ordinary, non-privileged sense. For example,
\texttt{tool cabinet}, \texttt{cleaning agent}, and
\texttt{trace the path} are treated as matches. Because the plural and
apostrophe-plus-\texttt{s} suffixes are independently optional, the
implementation may also match unusual strings such as \verb|agents's|;
this edge case does not alter the stated sample-level aggregation rule.
Overall, the loose criterion provides a conservative surface-form
leakage measure and is distinct from stricter semantic filtering
procedures.

\subsubsection{Sample-Level Leakage Indicator}

Let \(x_i\) denote the generated response for evaluation sample \(i\),
and let \(c_{i,w}\) be the number of regex matches for keyword \(w\) in
that response. The per-keyword occurrence count is
\begin{equation}
    c_{i,w}
    =
    \left|
        \operatorname{RegexFindAll}(x_i, r_w)
    \right|,
    \label{eq:keyword_occurrence_count}
\end{equation}
where \(r_w\) is the keyword-specific expression defined above.
These counts may be retained for diagnostic analysis, including
separate reporting of the numbers of \texttt{agent}, \texttt{trace},
and \texttt{tool} occurrences.

For the leakage-rate calculation, however, each response contributes
only a binary outcome. We define
\begin{equation}
    L_i
    =
    \mathbb{1}\!\left[
        \sum_{w \in \{\texttt{agent}, \texttt{trace}, \texttt{tool}\}}
        c_{i,w}
        > 0
    \right].
    \label{eq:sample_leakage_indicator}
\end{equation}

Accordingly, \(L_i = 1\) if the response contains at least one qualifying
occurrence of any target keyword, and \(L_i = 0\) otherwise. Multiple
occurrences of one keyword, or simultaneous occurrences of several
keywords, do not increase the contribution of that response beyond one.

\subsubsection{Dataset-Level Leakage Rate}

Given \(N\) evaluated samples, the privileged information leakage rate
is computed as
\begin{equation}
    \operatorname{PILR}
    =
    \frac{1}{N}\sum_{i=1}^{N} L_i
    =
    \frac{\#\{i : L_i = 1\}}{N}.
    \label{eq:privileged_information_leakage_rate}
\end{equation}

The denominator is the total number of evaluated samples, while the
numerator is the number of responses containing at least one qualifying
match. For example, a response containing both \texttt{agent} and
\texttt{tool} several times contributes one leaked sample to the
numerator, not multiple leaked samples. The metric may be multiplied
by \(100\) when reported as a percentage.

\subsubsection{Reference Implementation}

\begin{verbatim}
import re

LOOSE_LEAK_WORDS = {
    word: re.compile(rf"\b{word}s?(?:['\u2019]s)?\b", re.I)
    for word in ("agent", "trace", "tool")
}

ANY_LOOSE_WORD = re.compile(
    r"\b(?:agent|trace|tool)s?(?:['\u2019]s)?\b",
    re.I,
)

def keyword_counts(text: str) -> dict[str, int]:
    return {
        name: len(pattern.findall(text))
        for name, pattern in LOOSE_LEAK_WORDS.items()
    }

def sample_has_leakage(text: str) -> bool:
    return ANY_LOOSE_WORD.search(text) is not None

def privileged_information_leakage_rate(
    responses: list[str],
) -> float:
    return (
        sum(sample_has_leakage(text) for text in responses)
        / len(responses)
    )
\end{verbatim}

\subsection{What does SpatialOPSD learn?}
\label{app:analysis}

We compare the reasoning rollouts produced by the base model and the spatial-reasoning-trained model on the same 1,040 MindCube-tiny examples. The categories below are defined by explicit lexical markers. Within each category, the markers are non-overlapping, and the category count is the sum of its marker counts. To control for differences in response length, we report occurrences per 1,000 words. We additionally report \emph{rollout coverage}, defined as the percentage of examples whose corresponding rollout contains at least one marker from the category:
\begin{equation}
\operatorname{Coverage}(c)
= \frac{\#\{i:\text{rollout }i\text{ contains at least one marker in }c\}}{1040}\times 100\%.
\end{equation}

\begin{figure}[!hb]
    \centering
    \includegraphics[width=\linewidth]{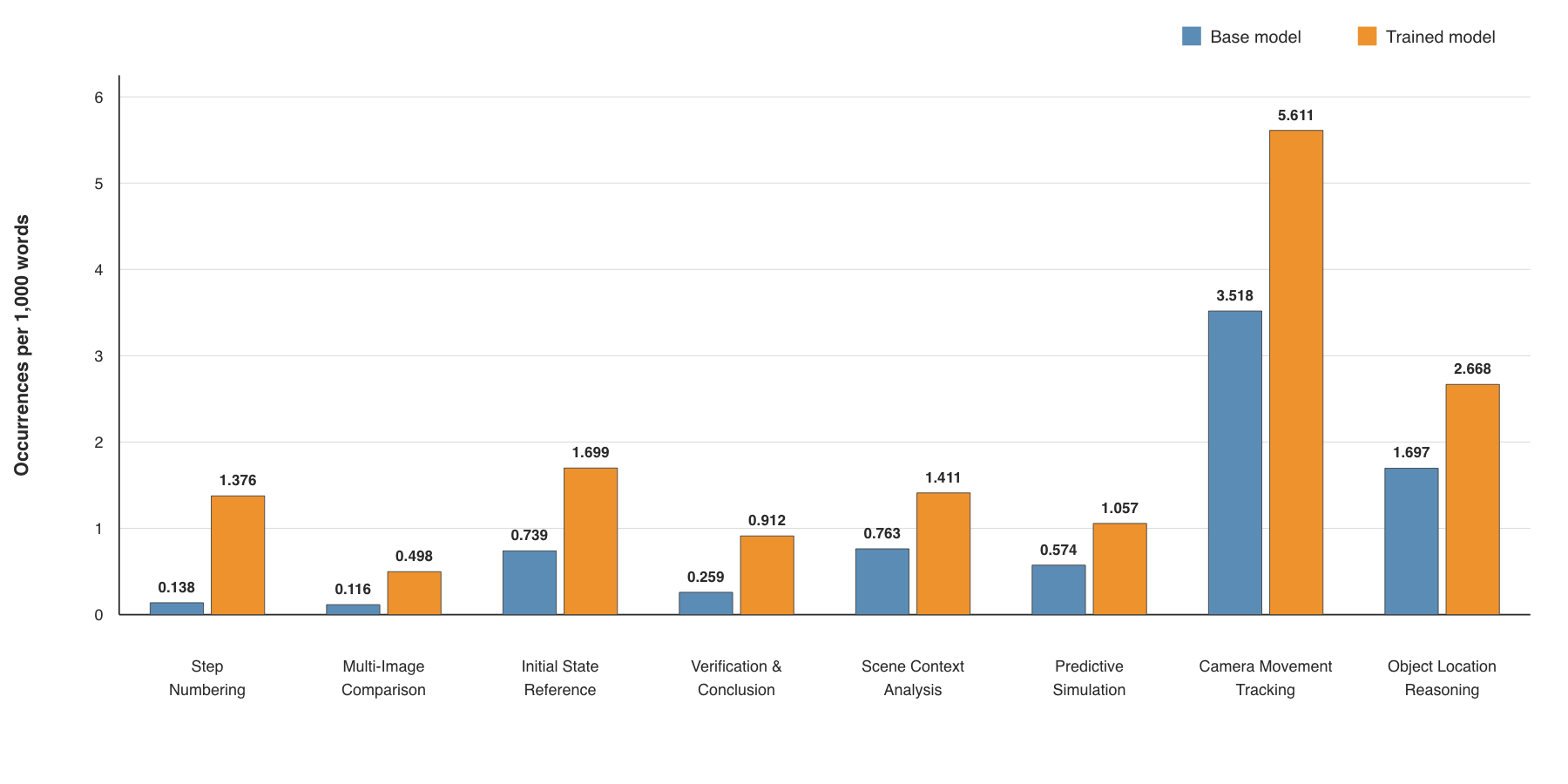}
    \caption{Category-level lexical frequency before and after spatial-reasoning training. Frequencies are normalized by rollout length and reported as occurrences per 1,000 words. The value above each bar gives the exact plotted frequency.}
    \label{fig:appendix-term-frequency}
\end{figure}

\begin{figure}[t]
    \centering
    \includegraphics[width=\linewidth]{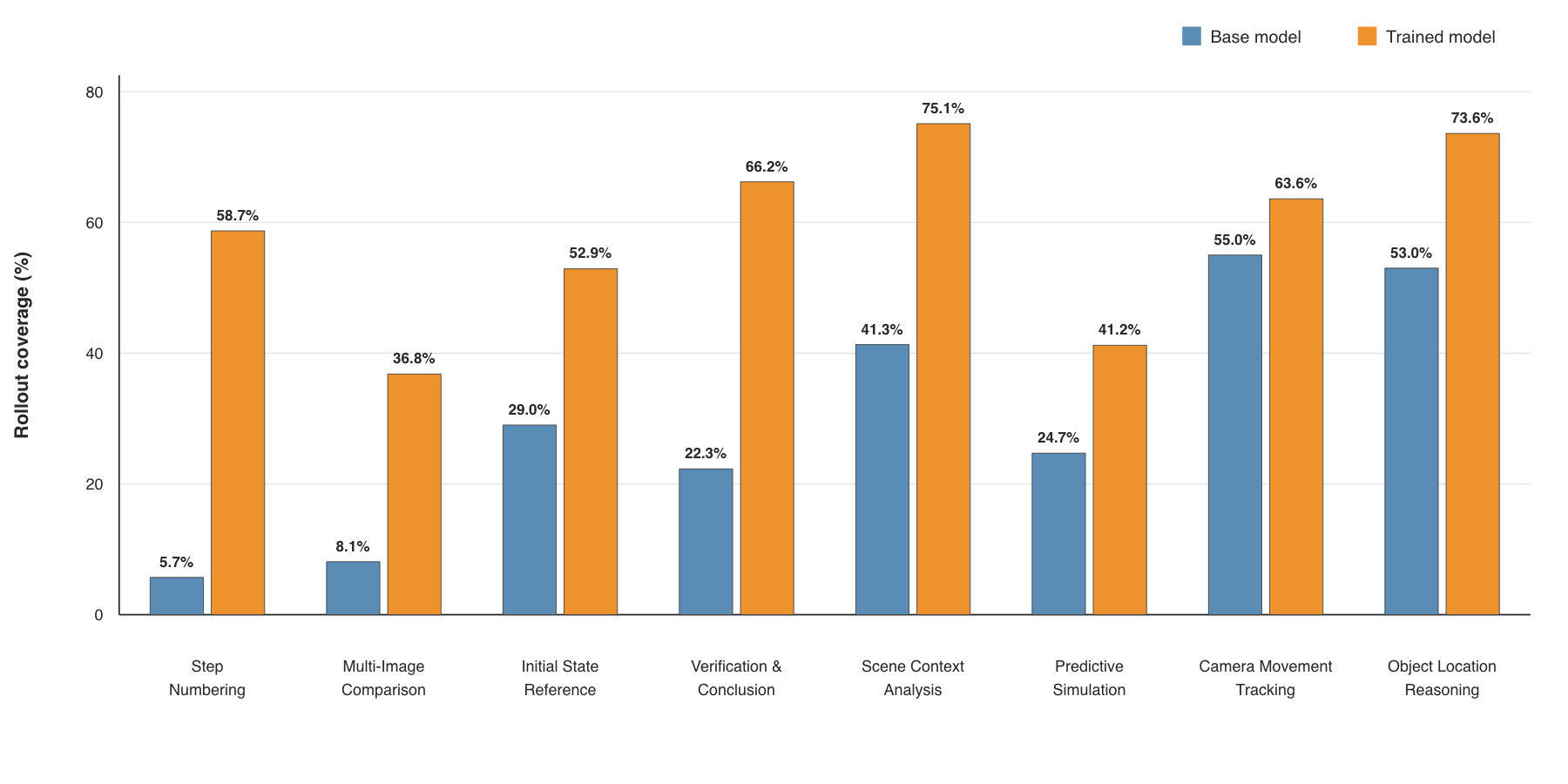}
    \caption{Rollout coverage before and after spatial-reasoning training. Coverage is the percentage of the 1,040 examples whose rollout contains at least one marker from the corresponding category. The value above each bar gives the exact plotted percentage.}
    \label{fig:appendix-rollout-coverage}
\end{figure}

Figures~\ref{fig:appendix-term-frequency} and~\ref{fig:appendix-rollout-coverage} compare the category-level lexical frequency and rollout coverage, respectively, before and after spatial-reasoning training. All eight categories become more prevalent after training under both measures. Camera Movement Tracking has the highest normalized frequency, increasing from 3.518 to 5.611 occurrences per 1,000 words. Step Numbering shows the largest relative frequency increase, from 0.138 to 1.376. In terms of coverage, the largest absolute increases occur for Step Numbering ($+53.0$ percentage points), Verification and Conclusion ($+43.9$ points), and Scene Context Analysis ($+33.8$ points). These measurements are descriptive lexical indicators of changes in the model's reasoning traces; by themselves, they should not be interpreted as causal evidence of improved reasoning. Table~\ref{tab:appendix-lexical-markers} lists the exact lexical markers used to define each category and compute its frequency and rollout coverage.

% \begin{itemize}
%     \item \textbf{Step Numbering:} \texttt{step 1}, \texttt{step 2}, and \texttt{step 3}.
%     \item \textbf{Multi-Image Comparison:} \texttt{compare}, \texttt{both images}, and \texttt{using the three}.
%     \item \textbf{Initial State Reference:} \texttt{original} and \texttt{initial}.
%     \item \textbf{Verification and Conclusion:} \texttt{understand}, \texttt{verify}, and \texttt{conclusion}.
%     \item \textbf{Scene Context Analysis:} \texttt{scene}, \texttt{context}, and \texttt{situation}.
%     \item \textbf{Predictive Simulation:} \texttt{if the camera}, \texttt{into view}, \texttt{would bring}, \texttt{would make}, \texttt{will bring}, \texttt{would move}, \texttt{brings you closer}, \texttt{would shift}, and \texttt{would appear}.
%     \item \textbf{Camera Movement Tracking:} \texttt{forward}, \texttt{camera moved}, and \texttt{trace}.
%     \item \textbf{Object Location Reasoning:} \texttt{position} and \texttt{located}.
% \end{itemize}

% \clearpage
\begin{table}[t!]
    \centering
    \caption{Lexical markers used to define each reasoning category.}
    \label{tab:appendix-lexical-markers}
    \small
    \renewcommand{\arraystretch}{1.15}
    \begin{tabular}{p{0.28\linewidth} p{0.66\linewidth}}
        \hline
        \textbf{Category} & \textbf{Lexical markers} \\
        \hline
        Step Numbering & \texttt{step 1}, \texttt{step 2}, \texttt{step 3} \\
        Multi-Image Comparison & \texttt{compare}, \texttt{both images}, \texttt{using the three} \\
        Initial State Reference & \texttt{original}, \texttt{initial} \\
        Verification and Conclusion & \texttt{understand}, \texttt{verify}, \texttt{conclusion} \\
        Scene Context Analysis & \texttt{scene}, \texttt{context}, \texttt{situation} \\
        Predictive Simulation & \texttt{if the camera}, \texttt{into view}, \texttt{would bring}, \texttt{would make}, \texttt{will bring}, \texttt{would move}, \texttt{brings you closer}, \texttt{would shift}, \texttt{would appear} \\
        Camera Movement Tracking & \texttt{forward}, \texttt{camera moved}, \texttt{trace} \\
        Object Location Reasoning & \texttt{position}, \texttt{located} \\
        \hline
    \end{tabular}
\end{table}